\documentclass[runningheads]{llncs}

\usepackage{eccv}

\usepackage{eccvabbrv}

\usepackage{booktabs}       %
\usepackage{amsfonts}       %
\usepackage{nicefrac}       %
\usepackage{microtype}      %
\usepackage{xcolor}         %
\usepackage{graphicx}
\usepackage{amsmath}
\usepackage{amssymb}
\usepackage{array}
\usepackage{multirow}
\usepackage{color}
\usepackage{colortbl}
\usepackage{framed}
\usepackage{bm}
\usepackage{xspace}
\usepackage{enumitem}
\usepackage{xparse}
\usepackage{listings}
\usepackage{url}
\usepackage{mathtools}
\usepackage{pifont}
\usepackage{mathtools}
\usepackage{lipsum}
\usepackage{adjustbox}
\usepackage{wrapfig}

\usepackage{siunitx} %
\usepackage{algorithm}%
\usepackage{algorithmicx}%
\usepackage{algpseudocode}
\usepackage{setspace}
\usepackage{comment}
\usepackage{makecell}

\definecolor{Gray}{gray}{0.2}
\definecolor{lightgray}{gray}{0.92}
\definecolor{TitleColor}{gray}{0.95}

\definecolor{blond}{rgb}{0.98, 0.94, 0.75}
\definecolor{LightCyan}{rgb}{0.88,0.95,1}
\definecolor{OurColor}{rgb}{0.886, 0.941, 0.851}

\def \ie {\emph{i.e.}}
\def \eg {\emph{e.g.}}

\newcommand{\tit}[1]{\smallbreak\noindent\textbf{#1.}}
\newcommand{\tinytit}[1]{\noindent\textbf{#1.}}

\newcommand{\ours}{Dress-EM\xspace}
\newcommand{\dataset}{Dress-ED\xspace}

\definecolor{my_green}{RGB}{51,102,0}
\definecolor{my_red}{RGB}{204, 0, 0}
\renewcommand{\checkmark}{\textcolor{my_green}{\ding{51}}} %
\newcommand{\crossmark}{\textcolor{my_red}{\ding{55}}} %

\usepackage[accsupp]{axessibility}

\usepackage[pagebackref,breaklinks,colorlinks,citecolor=eccvblue]{hyperref}

\usepackage{orcidlink}

\begin{document}

\title{\raisebox{-0.2cm}{\includegraphics[width=0.8cm]{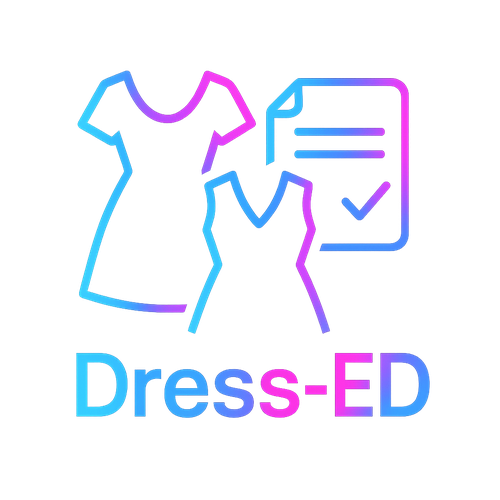}} \dataset: Instruction-Guided Editing for Virtual Try-On and Try-Off}

\titlerunning{Dress-ED: Instruction-Guided Editing for Virtual Try-On and Try-Off}

\author{
Davide Lobba\inst{1,3}$^{*}$\orcidlink{0009-0003-4042-7789} \and
Fulvio Sanguigni\inst{2,3}$^{*}$\orcidlink{0009-0002-9362-990X} \and
Bin Ren\inst{4}$^{\dagger}$\orcidlink{0000-0002-9790-1504} \and\\
Marcella Cornia\inst{2}\orcidlink{0000-0001-9640-9385} \and
Rita Cucchiara\inst{2}\orcidlink{0000-0002-2239-283X} \and
Nicu Sebe\inst{1}\orcidlink{0000-0002-6597-7248} 
}

\authorrunning{D.~Lobba, et al.}

\institute{
University of Trento, Italy \\
\email{\{davide.lobba, nicu.sebe\}@unitn.it} \and
University of Modena and Reggio Emilia, Italy \\
\email{\{fulvio.sanguigni, marcella.cornia, rita.cucchiara\}@unimore.it} \and
University of Pisa, Italy \and
MBZUAI, United Arab Emirates \\
\email{bin.ren@mbzuai.ac.ae}
\\\vspace{0.2cm}
\href{https://aimagelab.github.io/Dress-ED/}{\texttt{aimagelab.github.io/Dress-ED/}}
}

\maketitle

% \begingroup
% \renewcommand{\thefootnote}{\fnsymbol{footnote}}
% \footnotetext[1]{Equal contribution.}
% \quad
% $^{\dagger}$ Corresponding author.
% \endgroup
\begingroup
\renewcommand{\thefootnote}{}
\footnotetext{
$^{*}$ Equal contribution.
$^{\dagger}$ Corresponding author.
}
\endgroup

\begin{center}
\vspace{-0.4cm}
    \includegraphics[width=\linewidth]{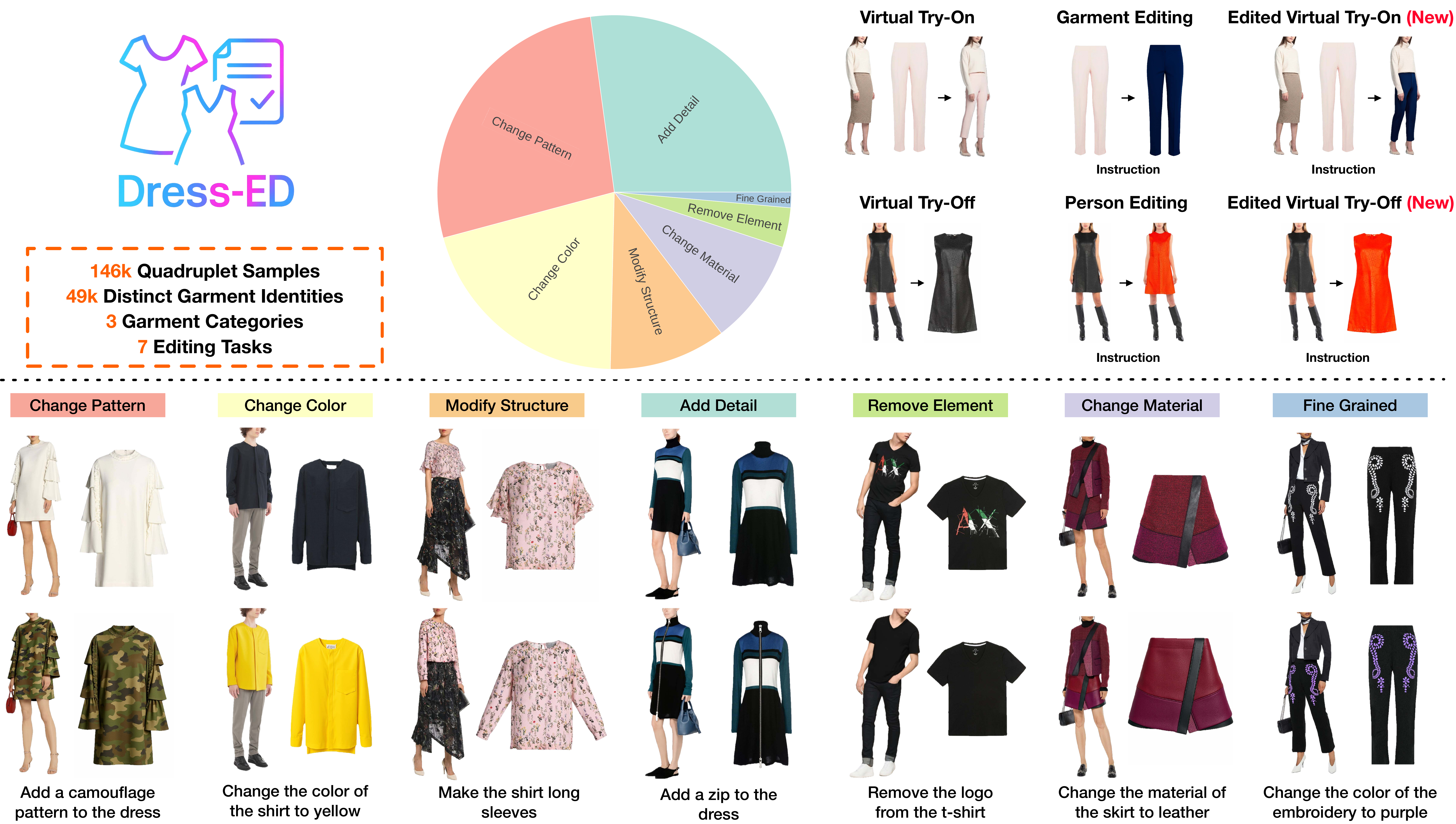}%
    \captionof{figure}{We propose the \textbf{Dress Editing Dataset} (\textbf{\dataset}), the first benchmark for instruction-driven virtual try-on and try-off with over 146k verified samples across seven editing types, including both appearance and structural modifications.
    }
    \label{fig:teaser}
    \vspace{-4mm}
\end{center}%

\begin{abstract}
  Recent advances in Virtual Try-On (VTON) and Virtual Try-Off (VTOFF) have greatly improved photo-realistic fashion synthesis and garment reconstruction. However, existing datasets remain static, lacking instruction-driven editing for controllable and interactive fashion generation. In this work, we introduce the \textbf{Dress Editing Dataset} (\textbf{\dataset}), the first large-scale benchmark that unifies VTON, VTOFF, and text-guided garment editing within a single framework. Each sample in \dataset includes an in-shop garment image, the corresponding person image wearing the garment, their edited counterparts, and a natural-language instruction of the desired modification. Built through a fully automated multimodal pipeline that integrates MLLM-based garment understanding, diffusion-based editing, and LLM-guided verification, \dataset comprises over 146k verified quadruplets spanning three garment categories and seven edit types, including both appearance (\eg, color, pattern, material) and structural (\eg, sleeve length, neckline) modifications. Based on this benchmark, we further propose a unified multimodal diffusion framework that jointly reasons over linguistic instructions and visual garment cues, serving as a strong baseline for instruction-driven VTON and VTOFF.
\end{abstract}

\sloppy
\section{Introduction}
\label{sec:intro}
Digital garment manipulation on human images has emerged as a central challenge at the intersection of computer vision, computer graphics, and fashion intelligence. Tasks such as Virtual Try-On (VTON)~\cite{han2018viton,morelli2022dresscode,li2024anyfit} and Virtual Try-Off (VTOFF)~\cite{lobba2025inverse,velioglu2024tryoffdiff} aim to bridge physical garments and digital humans through realistic cross-modal synthesis and reconstruction. Recent progress in diffusion-based generation~\cite{zhu2023tryondiffusion,baldrati2023multimodal,chong2025catvton} and Transformer architectures~\cite{kim2023stableviton,jiang2024fitdit,li2025ditvton} has achieved remarkable realism. Nevertheless, existing datasets remain largely static, containing fixed person-garment pairs without support for instruction-level control or semantic supervision, thereby constraining progress toward interactive and controllable fashion manipulation.

Beyond conventional VTON and VTOFF, emerging applications demand fine-grained garment editing, such as modifying color, texture, or silhouette via natural-language instructions. Achieving this requires precise alignment between garment semantics, human appearance, and linguistic intent. Yet, the existing efforts remain fragmented:
\textit{(i)} VTON datasets~\cite{han2018viton,choi2021vitonhd,morelli2022dresscode} focus on person–garment synthesis without editability; 
\textit{(ii)} instruction-driven datasets~\cite{brooks2023instructpix2pix,zhang2023magicbrush,hui2024hq,zhao2024ultraedit} support generic image editing but lack consistent garment–person correspondences; and \textit{(iii)} fashion editing approaches~\cite{zhu2024mmvtomultigarmentvirtual,yin2025editgarment,Huang_2025_T-Fit} tackle narrow sub-tasks without unified benchmarks. As a result, controllable, multimodal fashion editing remains underexplored.

Motivated by these limitations, in this work we propose a unified benchmark for instruction-driven fashion editing that jointly captures garment semantics, human appearance, and editing intent. To this end, we introduce the \textbf{Dress Editing Dataset} (\textbf{\dataset}), the first large-scale, systematically curated benchmark that integrates VTON and VTOFF tasks under a single instruction-based framework. Each sample in Dress-ED includes an in-shop garment image, the corresponding person image wearing the garment, their edited counterparts, and a natural-language instruction describing the desired modification (Fig.~\ref{fig:teaser}). The dataset is constructed through a fully automated multimodal pipeline. In particular, the pipeline first extracts structured garment attributes using Qwen3-VL~\cite{bai2025qwen3}, then synthesizes diverse editing instructions and applies diffusion-based editing via FLUX.2 Klein~\cite{flux-2-2025}. Finally, multi-stage quality verification with GPT-5~\cite{openaigpt-5} and a fine-tuned InternVL-3.5~\cite{wang2025internvl3} ensures semantic consistency and visual fidelity. The resulting dataset comprises over 146k verified samples spanning three garment categories and seven editing types, covering both \textit{appearance edits} (\eg, color, pattern, material) and \textit{structural modifications} (\eg, sleeve length, neckline, details). \dataset thus establishes a scalable and semantically grounded benchmark for controllable, instruction-driven fashion generation.

In addition to dataset construction, \dataset contributes along two key dimensions: \textit{benchmarking} and \textit{methodology}. It introduces two complementary benchmarks (\ie, instruction-driven VTON and VTOFF) under unified evaluation protocols assessing both visual realism and semantic edit accuracy. Building on this foundation, we propose a unified multimodal diffusion framework for instruction-driven editing in both try-on and try-off scenarios. The framework integrates a pretrained MLLM, a lightweight connector, and a DiT-based component~\cite{peebles2023scalable} to jointly reason over textual instructions and visual garment cues.
This design enables fine-grained, localized edits through token-level conditioning while preserving global semantic coherence, providing a strong and extensible baseline for future research in multimodal fashion understanding and generation.

\noindent\textbf{Contributions.} Our main contributions are threefold:
\begin{itemize}
\item We present Dress-ED, the first large-scale dataset that unifies VTON, VTOFF, and instruction-driven garment editing within a single, semantically consistent framework.
\item We develop a fully automated multimodal pipeline that combines MLLM-based garment understanding, diffusion-based synthesis, and LLM-guided verification, generating over 146k high-quality samples across diverse edit types.
\item We establish comprehensive benchmarks and a unified diffusion-based baseline for instruction-driven fashion editing in both VTON and VTOFF settings.
\end{itemize}

\section{Related Work}
\label{sec:related}
\tinytit{Virtual Try-On and Try-Off} VTON aims to synthesize realistic images of a target person wearing a given garment~\cite{bai2022single,cui2021dressing,fele2022c,ren2023cloth}, while VTOFF addresses the inverse process of recovering clean garment representations from dressed-person photos for digitization and editing. Early \textit{warping-based} methods~\cite{han2018viton,wang2018toward,chen2023size,xie2023gpvton,yan2023linking} deform garments to align with body shapes but often suffer from misaligned textures and ghosting artifacts. Recent \emph{warping-free} models~\cite{zhu2023tryondiffusion,morelli2023ladi,baldrati2023multimodal,chong2025catvton} remove explicit deformation by using attention-based conditioning, achieving higher realism though sometimes losing fine texture fidelity. To mitigate this, approaches based on diffusion-based architectures~\cite{kim2023stableviton,jiang2024fitdit,zhu2024mmvtomultigarmentvirtual,li2025ditvton} enhance garment details, while conditioning modules such as LOTS~\cite{girella2025lotsfashionmulticonditioningimage} and LEFFA~\cite{zhou2024learningflowfieldsattention} improve structural consistency. 
In parallel, diffusion-based VTOFF models~\cite{velioglu2024tryoffdiff,xarchakos2024tryoffanyone,velioglu2025mgt,lobba2025inverse} advance garment disentanglement from occlusions and complex poses across multiple categories. More unified frameworks like Voost~\cite{lee2025voost}, One-Model-for-All~\cite{liu2025one}, and Any2AnyTryon~\cite{guo2025any2anytryonleveragingadaptiveposition} attempt to integrate both VTON and VTOFF within a single model. However, the progress of such unified modeling remains limited by the lack of large-scale datasets that jointly support person-centric synthesis and garment-centric reconstruction with fine-grained correspondence. Motivated by this gap, we introduce a large-scale dataset that unifies both VTON and VTOFF tasks, enabling detailed garment editing and realistic human-garment interaction beyond prior approaches.

\begin{figure*}[t]
    \centering
    \includegraphics[width=\linewidth]{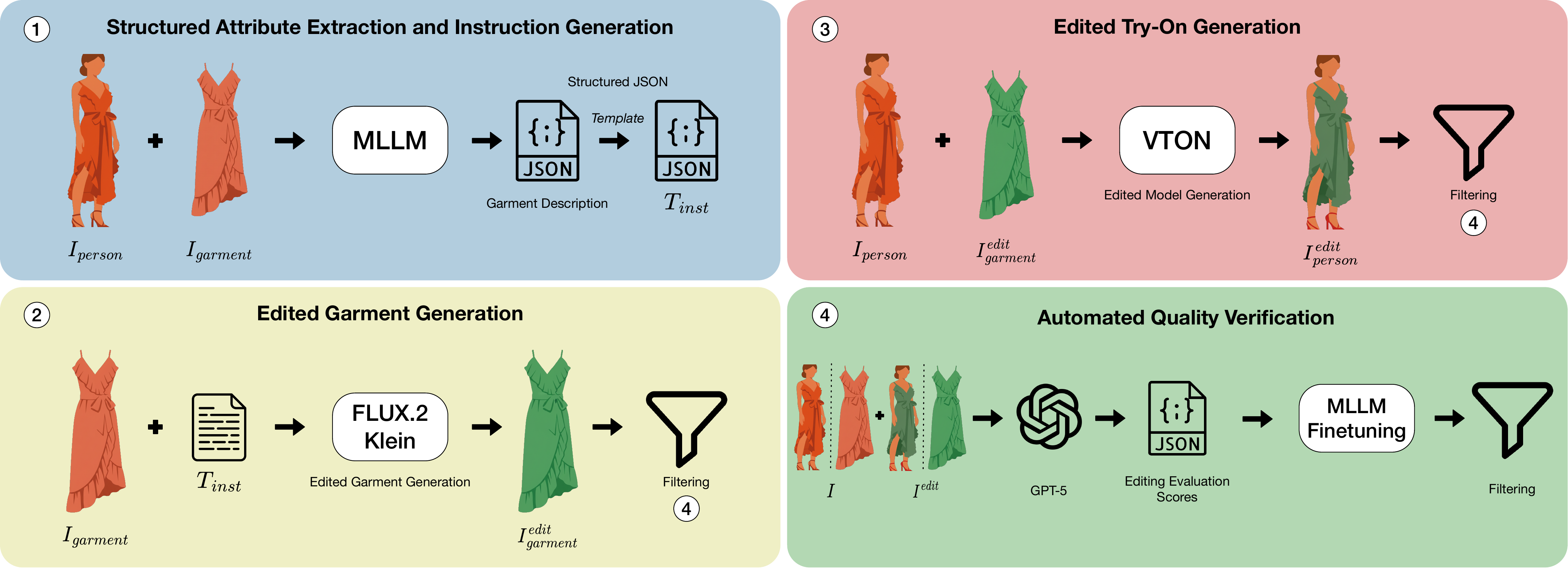}
    \caption{\textbf{Overview of the \dataset curation pipeline.} 
    Starting from Dress Code~\cite{morelli2022dresscode}, (i) structured garment attributes are extracted with Qwen3-VL and used to generate natural-language edit instructions, (ii) edited in-shop garments are synthesized via FLUX.2 Klein, (iii) the corresponding edited try-on images are generated using FitDiT, and (iv) all results are validated through a version of InternVL-3.5, fine-tuned with samples annotated using GPT-5~\cite{openaigpt-5}, to ensure semantic and visual consistency.
    } %
    \label{fig:dataset_pipeline}
    \vspace{-0.4cm}
\end{figure*}

\tit{General-Purpose Editing} 
Image editing has become an active research area, enabling flexible and fine-grained manipulation of image content. Existing approaches can be broadly categorized into several types. Finetuning-based methods~\cite{brooks2023instructpix2pix,zhang2023magicbrush,fu2023guidingmgie,zhang2024hive,wei2024omniedit,sheynin2024emu,yang2024editworld,liu2025step1x-edit,yu2024anyedit} adapt pretrained generative models using real or synthetic supervision. Some studies~\cite{fu2023guidingmgie,liu2025step1x-edit} further incorporate MLLM guidance to inject richer multimodal cues. Other research lines explore zero-shot editing via noise or flow inversion~\cite{garibi2024renoise,patel2024flowchef,jiao2025uniedit,rout2024rfinversion} and probing-based analysis to identify edit-receptive components~\cite{avrahami2025stable}. However, these general-purpose methods often underperform in fashion scenarios, where garments exhibit complex geometry, layered structure, and fine-grained textures that require precise spatial and semantic control.
Recently, foundation models such as FLUX.2 Klein~\cite{flux-2-2025} and Qwen-Image-Edit~\cite{wu2025qwen} have advanced instruction-based image-to-image editing, supporting diverse manipulation tasks. Yet, these models remain limited when the input and target belong to different domains (\eg, person-to-garment or garment-to-person transformations).

\tit{Fashion-Oriented Editing}
In the VTON setting, several methods have explored instruction-based editing as an auxiliary task. M\&M VTO~\cite{zhu2024mmvtomultigarmentvirtual} and HieraFashDiff~\cite{xie2024hierarchical} demonstrate qualitative results on simple edits such as rolled sleeves. However, these works neither provide quantitative benchmarking nor address more challenging edits involving complex structures or patterns. In the broader fashion editing domain, recent efforts focus on either person-to-person~\cite{Huang_2025_T-Fit} or garment-to-garment~\cite{yin2025editgarment} transformations. T-FIT~\cite{Huang_2025_T-Fit} guides the model using concept vectors derived from CLIP text embeddings and fine-tunes it via LoRA~\cite{hu2022lora} to preserve non-instructional visual attributes. EditGarment~\cite{yin2025editgarment} constructs synthetic pairs of original and target garments with corresponding textual captions and trains an InstructPix2Pix~\cite{brooks2023instructpix2pix} model for garment-level editing. Despite these advances, no existing approach can jointly perform VTON or VTOFF and instruction-driven editing within a unified framework. Moreover, the lack of a reliable benchmark with ground-truth supervision limits consistent evaluation of edit quality and semantic fidelity. Our proposed \dataset bridges these gaps by unifying VTON, VTOFF, and instruction-based editing for dedicated model training and evaluation.

\section{Dress Editing Dataset (\dataset)}
\label{sec:dataset}

We introduce the \textbf{Dress Editing Dataset} (\textbf{\dataset}), a large-scale benchmark unifying instruction-driven fashion editing for both VTON and VTOFF. \dataset extends Dress Code~\cite{morelli2022dresscode} by adding natural-language editing instructions and corresponding edited images, aligning textual intent with visual garment semantics. Each sample supports \textit{appearance} edits (\eg, color, pattern, material) and \textit{structural} edits (\eg, sleeve length, neckline, details), enabling fine-grained and controllable garment manipulation. This unified formulation provides a comprehensive benchmark for evaluating and training instruction-guided fashion editing models.

\begin{algorithm}[t]
\scriptsize
\setstretch{1.1} 
\caption{Automated Multimodal Data Generation.}
\label{alg:data_pipeline}
\begin{algorithmic}[1]
    \Require Paired samples $\{(I_{\text{garment}}, I_{\text{person}})\}$ from Dress Code~\cite{morelli2022dresscode}
    \Ensure Verified quadruplets $\mathcal{S}^*=\{(I_{\text{garment}}, I_{\text{person}}, I_{\text{garment}}^{\text{edit}}, I_{\text{person}}^{\text{edit}}, T_{\text{inst}})\}$

    \Statex \textbf{Stage 1: Attribute Extraction and Instruction Generation}
    \State $\mathcal{A}_g \gets \text{Qwen3-VL}(I_{\text{garment}}, I_{\text{person}})$ \Comment{structured garment attributes}
    \State $T_{\text{inst}} \gets f_{\text{temp}}(\mathcal{A}_g)$ \Comment{template-based edit instruction}

    \Statex \textbf{Stage 2: Diffusion-Based Garment Editing}
    \State $I_{\text{garment}}^{\text{edit}} \gets \text{FLUX.2 Klein}(I_{\text{garment}}, T_{\text{inst}})$ \Comment{apply appearance/structure edit}

    \Statex \textbf{Stage 3: Try-On Image Generation}
    \State $I_{\text{person}}^{\text{edit}} \gets \text{FitDiT}(I_{\text{person}}, I_{\text{garment}}^{\text{edit}})$ \Comment{render edited garment on person}

    \Statex \textbf{Stage 4: LLM-Guided Quality Verification}
    \State $s \gets \text{GPT-5Judge}(I, I^{\text{edit}}, T_{\text{inst}})$ \Comment{adherence \& quality scoring}
    \State $f_{\text{verifier}} \gets \text{Finetune}(\text{InternVL-3.5},\{(I,I^{\text{edit}},T_{\text{inst}},s)\}_{\sim5\text{k}})$
    \State $\mathcal{S}^* \gets \{(I_{\text{garment}},I_{\text{person}},I_{\text{garment}}^{\text{edit}},I_{\text{person}}^{\text{edit}},T_{\text{inst}}) \mid f_{\text{verifier}}(I,I^{\text{edit}},T_{\text{inst}})> t\}$

    \State \Return $\mathcal{S}^*$ \hfill {\color{gray}// $\sim$146k verified samples, 3 categories, 7 edit types}
\end{algorithmic}
\end{algorithm}

\subsection{Automated Dataset Generation Pipeline}
\label{subsec:pipline}
We construct \dataset through a fully automated multimodal pipeline, summarized in Alg.~\ref{alg:data_pipeline} and illustrated in Fig.~\ref{fig:dataset_pipeline}, which converts static garment–person pairs into instruction-driven editing samples.
Built upon the Dress Code dataset~\cite{morelli2022dresscode}, which contains paired images ($I_{\text{person}}, I_{\text{garment}}$) across three representative clothing types (\ie, upper-body, lower-body, and dresses), the pipeline integrates multimodal understanding, 
diffusion-based generation, 
and LLM-guided verification to produce coherent instruction-grounded samples.
Each finalized instance in \dataset is formulated as:
\begin{equation}
\label{eq:dataset_sample}
\mathcal{S} = \left( I_{\text{garment}},\, I_{\text{person}},\, I_{\text{garment}}^{\text{edit}},\, I_{\text{person}}^{\text{edit}},\, T_{\text{inst}} \right),
\end{equation}
where ($I_{\text{person}}, I_{\text{garment}}$) are the original inputs, 
$I_{\text{garment}}^{\text{edit}}$ and $I_{\text{person}}^{\text{edit}}$ are their edited counterparts, and 
$T_{\text{inst}}$ denotes the natural-language instruction.

\tit{Stage 1: Structured Attribute Extraction and Instruction Generation} 
Given the in-shop garment $I_{\text{garment}}$ and the corresponding on-model image $I_{\text{person}}$, 
we employ Qwen3-VL~\cite{bai2025qwen3} to extract structured garment semantics:
\begin{equation}
A_g = f_{\text{MLLM}}(I_{\text{garment}}, I_{\text{person}}),
\end{equation}
where $A_g$ is a structured JSON with both descriptive text and explicit attributes (\eg, color, pattern, material, neckline, sleeve length).  
Natural-language edit instructions are then synthesized via rule-based templates:
\begin{equation}
T_{\text{inst}} = f_{\text{temp}}(A_g),
\end{equation}
covering two main categories, \ie, \textit{appearance edits} (color, pattern, material) and \textit{structural edits} (adding, removing, reshaping parts). 
For example, a ``blue cotton shirt with short sleeves'' may yield  ``change the sleeve length to long'' or  ``make the shirt yellow'', forming $(I_{\text{garment}}, T_{\text{inst}})$ pairs.

\tit{Stage 2: Edited Garment Generation} 
Given the $(I_{\text{garment}}, T_{\text{inst}})$ pair, 
we employ the diffusion-based editor FLUX.2 Klein~\cite{flux-2-2025} to synthesize edited in-shop garments:
\begin{equation}
I_{\text{garment}}^{\text{edit}} = f_{\text{diff}}(I_{\text{garment}}, T_{\text{inst}}),
\end{equation}
yielding approximately 300k edited samples across categories, which are subsequently refined through automated quality verification to ensure realism and instruction adherence.

\tit{Stage 3: Edited Try-On Generation} 
We take $I_{\text{person}}$ and the edited garment $I_{\text{garment}}^{\text{edit}}$, and employ the state-of-the-art VTON model FitDiT~\cite{jiang2024fitdit} to generate its try-on result:
\begin{equation}
I_{\text{person}}^{\text{edit}} = f_{\text{VTON}}(I_{\text{person}}, I_{\text{garment}}^{\text{edit}}),
\end{equation}
where the person is realistically rendered wearing the modified garment. 
The generated try-on images are subsequently validated through the same automated quality verification pipeline to ensure semantic consistency and visual fidelity.

\begin{table*}[t]
    \centering
    \caption{%
    Comparison of fashion and general-purpose editing datasets. While existing datasets lack instruction-guided garment-person correspondences, \textbf{\dataset} introduces 146k verified samples enabling, for the first time, instruction-driven VTON and VTOFF.
    }
    \vspace{-0.15cm}
    \label{tab:dataset_comparison_fashion}
    \setlength{\tabcolsep}{.35em}
    \resizebox{\linewidth}{!}{
    \begin{tabular}{lc ccccl}
    \toprule
    \textbf{Dataset} & & \textbf{\# Samples} & \makecell{\textbf{\# Editing} \\ \textbf{Types}} & \makecell{\textbf{Automatic} \\ \textbf{Generated}} & \makecell{\textbf{MLLM} \\ \textbf{Filtering}} & \textbf{Supported Tasks} \\
    \midrule
    \rowcolor{magenta!5}
    \multicolumn{7}{l}{\textit{General-Purpose Editing Datasets}}\\
    MagicBrush~\cite{zhang2023magicbrush} & & 10,388 & 7 & \crossmark & \crossmark &  Visual Attributes Editing \\
    RefEdit~\cite{pathiraja2025refedit} & & ~20,000 & 5 & \checkmark & \crossmark &  Visual Attributes Editing \\
    UltraEdit~\cite{zhao2024ultraedit} & & 4,108,262 & 9 & \checkmark & \crossmark &  \begin{tabular}[c]{@{}l@{}} Visual Attributes, Style,\\ Background Editing \end{tabular}  \\
    HQ-Edit~\cite{hui2024hq}  & & ~450,000 & 7 & \checkmark & \checkmark &  Visual Attributes and Style Editing\\
    AnyEdit~\cite{yu2024anyedit} & & ~2,500,000 & 25 & \checkmark & \checkmark & \begin{tabular}[c]{@{}l@{}} Visual Attributes, Camera Movements,\\ Scene Editing \end{tabular} \\
    ImgEdit~\cite{ye2025imgedit} & & ~1,200,000 & 11 & \checkmark & \checkmark &  Visual Attributes and Garment Editing  \\
    \midrule
    \rowcolor{magenta!5}
    \multicolumn{7}{l}{\textit{Fashion-Oriented Datasets}}\\
    DeepFashion2~\cite{ge2019deepfashion2versatilebenchmarkdetection} & & 873,234 & - & - & - & \begin{tabular}[c]{@{}l@{}} Cloth Retrieval, Segmentation,\\ Detection, Pose Estimation \end{tabular} \\
    Street TryOn~\cite{cui2024street} & & 14,453 & - & - & - & Person-to-Person VTON \\
    VITON-HD~\cite{choi2021vitonhd} & & 13,679 & - & - & - &  VTON, VTOFF  \\
    Dress Code~\cite{morelli2022dresscode} & & 53,792 & - & - & - &  VTON, VTOFF  \\
    \midrule
    EditGarment~\cite{yin2025editgarment} & & 20,596 & 6 & \checkmark & \checkmark &  Instruction-Driven Garment Editing \\
    \rowcolor{magenta!15}
    \textbf{\dataset (Ours)} & & 146,460 & 7 & \checkmark & \checkmark & \begin{tabular}[c]{@{}l@{}}Instruction-Driven VTON and VTOFF, \\ Person Editing, Garment Editing,\\ VTON, VTOFF \end{tabular}\\
    \bottomrule
    \end{tabular}
}
\vspace{-0.3cm}
\end{table*}

\tit{Stage 4: Automated Quality Verification}
To ensure semantic correctness and visual fidelity, we adopt a three-step verification pipeline.  
First, each triplet $(I, I^{\text{edit}}, T_{\text{inst}})$ is evaluated by GPT-5~\cite{openaigpt-5}, which outputs an edit score $s \!\in\! [0,100]$ assessing instruction adherence, content preservation, and realism.  
Second, a subset of $\sim$5k annotated samples is used to fine-tune InternVL-3.5~\cite{wang2025internvl3}, distilling GPT-5’s scoring capability into a scalable multimodal evaluator:
\begin{equation}
f_{\text{verifier}} \!\leftarrow\! \text{Finetune}(\text{InternVL-3.5}, \{(I^{\text{edit}}, T_{\text{inst}}, s)\}),
\end{equation}
where $I^{\text{edit}}$ denotes either $I_\text{garment}^{\text{edit}}$ or $I_\text{person}^{\text{edit}}$ depending on the verification target.

Finally, $f_{\text{InternVL}}$ is applied to all edited garments and try-on results, and samples with predicted scores below a threshold $t$ are filtered out:
\begin{align}
\mathcal{S}^{*} = \{ &(I_\text{garment}, I_\text{person}, I_\text{garment}^{\text{edit}}, I_\text{person}^{\text{edit}}, T_{\text{inst}}) \mid \nonumber & f_{\text{verifier}}(I^{\text{edit}}, T_{\text{inst}}) > t \},
\end{align}
where $t$ is set to 80 after manual verification, retaining only samples for which both $I_\text{garment}^{\text{edit}}$ and $I_\text{person}^{\text{edit}}$ obtain a score above this threshold.

This process yields $\sim$146k verified high-quality samples across all garment categories, forming the final \dataset.

\subsection{Dataset Statistics}
\label{subsec:statistic}
\begin{wraptable}{r}{0.52\linewidth}
\vspace{-0.5cm}
    \centering
    \caption{Sample distribution by editing type.}
    \label{tab:edit_stats}
    \vspace{0.1cm}
    \resizebox{0.98\linewidth}{!}{
    \begin{tabular}{lrr}
    \toprule
    \textbf{Edit Type} & \textbf{Count} & \textbf{Proportion (\%)} \\
    \midrule
    Add Detail & 39,776 & 27 \\
    Change Pattern & 39,559 & 27 \\
    Change Color & 29,983 & 20 \\
    Modify Structure & 15,670 & 11 \\
    Change Material & 14,091 & 10 \\
    Remove Element & 5,305 & 4 \\
    Fine-Grained & 2,076 & 1 \\
    \midrule
    \textbf{Total} & 146,460 & 100.0 \\
    \bottomrule
    \end{tabular}}
    \vspace{-0.6cm}
\end{wraptable}
The final \dataset comprises 146,460 visual quadruplets 
$(I_{\text{garment}}, I_{\text{person}}, I_{\text{garment}}^{\text{edit}}, I_{\text{person}}^{\text{edit}})$ 
paired with natural-language instructions $T_\text{inst}$, covering three representative garment categories: dresses (80,865 samples), upper-body (45,567 samples), and lower-body (20,028 samples), encompassing 49,664 distinct garment identities and 6,073 unique editing instructions. 
We follow the same train/test protocol as Dress Code~\cite{morelli2022dresscode}, 
splitting the data into 132,201 training and 14,259 test samples with no overlap between partitions. All images are standardized to a resolution of 768$\times$1024.

\tit{Editing Distribution}
We define seven editing types, grouped into two macro-categories: \textbf{appearance edits}, comprising \textit{Change Color}, \textit{Change Pattern}, \textit{Change Material}, and \textit{Fine-Grained}, which modify the visual properties of the garment; and 
\textbf{structural edits}, comprising \textit{Add Detail}, \textit{Remove Element}, and \textit{Modify Structure}, which alter its geometry or composition by adding, removing, or reshaping garment parts.
Overall, the dataset exhibits substantial diversity across garment categories and edit types, establishing a comprehensive benchmark for instruction-driven VTON and VTOFF research. Table~\ref{tab:edit_stats} provides an overview of the editing instruction statistics while a visual taxonomy of the edits is presented in the supplementary material.

\subsection{Dataset Comparison}
Existing fashion datasets~\cite{ge2019deepfashion2versatilebenchmarkdetection,lepage2023lrvs, cui2024street, choi2021vitonhd, morelli2022dresscode} primarily target either VTON or VTOFF, but none include edited counterparts or support instruction-driven transformations. For instance, VITON-HD~\cite{choi2021vitonhd} and Dress Code~\cite{morelli2022dresscode} contain real garment–person pairs but lack text-guided edits. At the same time, general-purpose editing datasets~\cite{pathiraja2025refedit,zhao2024ultraedit,hui2024hq,zhang2023magicbrush,brooks2023instructpix2pix,yang2025complexedit} are built upon data triplets including the editing instruction, but lack domain-specific pairs of garment-person data that suit VTON and VTOFF. EditGarment~~\cite{yin2025editgarment} collects a dataset of garment-only editing pairs, while ImgEdit~\cite{ye2025imgedit} contains garment data as a subsection of their dataset, lacking structured garment-person association. In contrast, our dataset introduces a novel visual quadruplet structure $(I_{\text{garment}}, I_{\text{person}}, I_{\text{garment}}^{\text{edit}}, I_{\text{person}}^{\text{edit}})$ paired with a natural-language instruction $T_{\text{inst}}$, where each edited garment and corresponding try-on image are coherently generated and verified. This formulation enables a range of instruction-driven fashion editing tasks, including edited VTON, edited VTOFF, person editing, and garment editing, while remaining compatible with standard VTON and VTOFF setups. A detailed comparison between \dataset and existing fashion and general-purpose datasets is presented in Table~\ref{tab:dataset_comparison_fashion}.

\subsection{Benchmark Tasks and Evaluation}
\label{subsec:benchmarking}
While \dataset can support a variety of instruction-driven fashion editing tasks, in this work we focus on two core benchmarks that reflect its primary objectives, \ie, \textit{instruction-driven VTON} and \textit{instruction-driven VTOFF}. Each benchmark targets a specific aspect of instruction-based fashion editing, either the correct rendering of edited garments on people or the accurate generation of these garments in a catalog style. To evaluate both visual realism and semantic edit correctness, we introduce a dual evaluation protocol, detailed below.

\tit{Instruction-Driven Virtual Try-On}
This benchmark evaluates the ability of a model to generate a realistic edited person image $I_{\text{person}}^{\text{edit}}$ conditioned on an edit instruction $T_{\text{inst}}$ and garment context. Following standard VTON literature~\cite{morelli2022dresscode,choi2021vitonhd}, two complementary configurations are considered.  
In the \textit{paired setting}, the model receives $(I_{\text{person}}, I_{\text{garment}}, T_{\text{inst}})$ and must generate $I_{\text{person}}^{\text{edit}}$ where the same garment worn by the person is modified according to $T_{\text{inst}}$. This configuration measures localized editing ability while preserving identity, pose, and background.  
In the \textit{unpaired setting}, the model is provided with $(I_{\text{person}}, I_{\text{garment}}', T_{\text{inst}})$, where $I_{\text{garment}}'$ refers to a different garment image. The goal is to synthesize the person wearing the garment $I_{\text{garment}}'$ edited as described by $T_{\text{inst}}$, thereby assessing practical generalization to real-world virtual shopping scenarios.

\tit{Instruction-Driven Virtual Try-Off}
This benchmark focuses on garment-level editing in the catalog domain. Given a person image $I_{\text{person}}$ and a textual instruction $T_{\text{inst}}$, the model must generate an edited catalog garment image $I_{\text{garment}}^{\text{edit}}$. The instruction specifies both the garment identity and the intended modification (\eg, ``make the skirt red''), and the target supervision is the corresponding edited in-shop image provided in \dataset. This task isolates the capacity to semantically manipulate garment appearance and structure independent of person-specific context.

\tit{On Using Real Images as Ground-truth Data}
For rigorous quantitative evaluation, the test phase employs an inverse-editing protocol. The model is tasked with reconstructing the original, unedited person image $I_{\text{person}}$ from its edited counterpart $I^{\text{edit}}_{\text{person}}$ using an inverse instruction $T^{r}_{\text{inst}}$. This guarantees that reference-based metrics are calculated strictly against real-world ground truth, accurately measuring localized editing ability while preserving identity, pose, and background without synthetic bias.
We adopt this evaluation setting for all of our tasks (VTON paired, unpaired, and VTOFF).

\tit{Evaluation Metrics}
We evaluate all benchmarks using two complementary metric families that jointly quantify visual fidelity and instruction adherence. 
\begin{itemize}
\item \textbf{Visual Fidelity.}  We employ standard perceptual and distributional metrics, including FID~\cite{parmar2021cleanfid}, KID~\cite{binkowski2018demystifying}, SSIM~\cite{wang2004ssim}, LPIPS~\cite{zhang2018perceptual}, and DISTS~\cite{Ding_2020}, computed independently for both VTON and VTOFF outputs.
\item \textbf{Edit Correctness.} %
To evaluate semantic alignment with the editing instruction, we compute DINO-I~\cite{oquab2024dinov2learningrobustvisual} similarity between the generated image and the ground-truth image to measure content-level consistency.
\end{itemize}

\tit{Benchmark Protocol}
All evaluations are conducted on the held-out test split (14,259 samples), with no overlap of garment identities across splits. For VTON, both paired and unpaired settings are reported separately. This unified protocol provides a standardized basis for evaluating instruction-driven fashion editing across VTON and VTOFF paradigms.

\section{Proposed Method Towards Editing}
\label{sec:method}
\begin{figure}[t]
    \centering
    \includegraphics[width=0.9\linewidth]{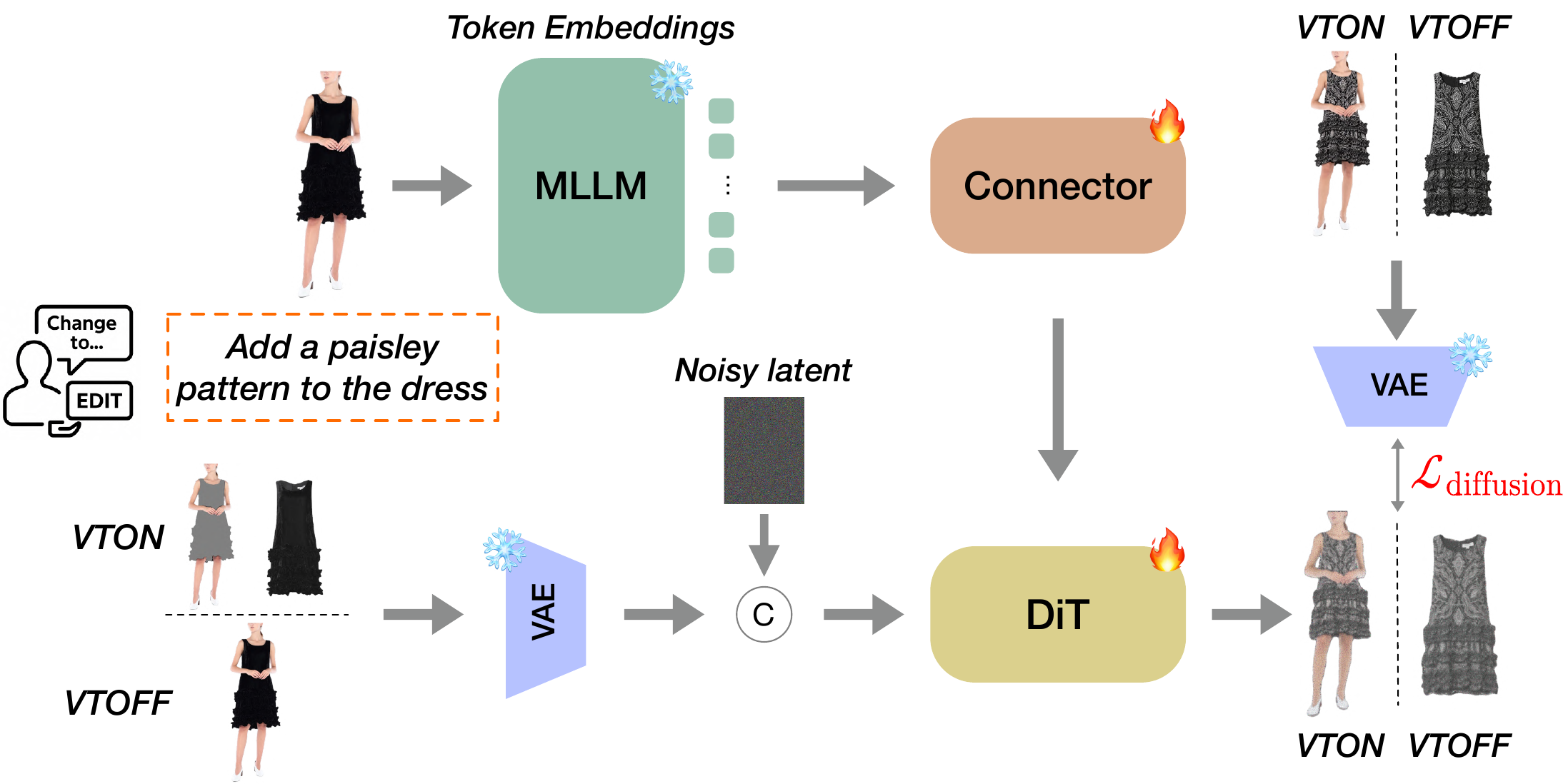}
    \vspace{-0.2cm}
    \caption{Overview of the proposed \ours architecture for instruction-driven fashion editing.}
    \label{fig:method}
    \vspace{-0.4cm}
\end{figure}

\vspace{-3pt}
We propose a unified multimodal diffusion architecture for instruction-driven editing, termed as \textbf{\ours} (\textbf{Dress Editing Model}) and applicable to both VTON and VTOFF tasks. 
Our framework unifies a pretrained multimodal large language model (MLLM), a lightweight connector, and a diffusion transformer (DiT) into a cohesive architecture, as illustrated in Fig.~\ref{fig:method}.
The only distinction between the two tasks lies in the inputs provided to the VAE encoder: 
for VTOFF, the input is the original person image, while 
for VTON, the encoder receives both the masked person and garment features, depending on the paired or unpaired setting.

\tit{Architecture Overview}
Given a visual reference $I$, the image of the original person, and a text instruction $t$, the MLLM (\ie, InternVL-3.5~\cite{wang2025internvl3}) jointly encodes the multimodal context through a single forward pass. This produces a sequence of fused token embeddings $\bm{E}_{\mathrm{MLLM}} = \{e_1, \ldots, e_N\}$ capturing fine-grained cross-modal relationships between the garment, the person, and the desired modification.
To isolate semantically relevant tokens, we retain only those aligned with the instruction and the image, discarding system prefix and irrelevant embeddings.

\tit{Connector}
The connector is a lightweight Transformer component that bridges the MLLM feature space and the DiT conditioning space. Specifically, it compresses $\bm{E}_{\mathrm{MLLM}}$ into a multimodal latent representation $\bm{z}_c=f_{conn}(\bm{E}_{\mathrm{MLLM}})$ and projects the mean pooled token embedding through a linear layer to produce a global guidance vector $\bm{g} \in \mathbb{R}^{2048}$ and a Token Refinement Pathway $z_c \in \mathbb{R}^{77 \times 4096}$.

\tit{Global Guidance Pathway ($g$)}
This pathway extracts a high-level semantic summary of the multimodal context produced by the MLLM. 
The connector aggregates the MLLM tokens via masked mean pooling to obtain a single context representation, 
which is then normalized and projected through a linear layer to produce a 2048-dimensional global guidance vector $g$. 

\tit{Token Refinement Pathway ($z_c$)}
This pathway preserves fine-grained multimodal information by refining the full sequence of MLLM tokens. The tokens are processed by a lightweight stack of Transformer blocks that resolve cross-modal interactions and produce a refined multimodal embedding $z_c \in \mathbb{R}^{77 \times 4096}$.

\tit{Diffusion Transformer}
Following the DiT paradigm~\cite{peebles2023scalable}, the edited image is generated in the latent space of a pretrained VAE. 
The DiT receives two complementary conditioning sources: \textit{(i)} the multimodal embedding $\bm{z}_c$ and a global guidance vector $\bm{g}$ produced by the MLLM and connector, which encode the semantic action of the edit, and \textit{(ii)} the VAE latent features $\bm{z}_{\mathrm{VAE}}$, which capture the spatial and structural details of the visual reference. For the VTON task, the $\bm{z}_{\mathrm{VAE}}$ is the concatenation along the width dimension of the latents of the masked person and the latents of the garment. Instead, for the VTOFF task, it is the latent of the garment. These signals allow the DiT to understand what to modify and where to apply it jointly.
Formally, the denoising process is defined as 
\begin{equation}
    \hat{\bm{x}}_{t-1} = \mathrm{DiT}(\bm{x}_t, \bm{z}_{\mathrm{VAE}}, \bm{z}_c, \bm{g}, t),
\end{equation}
where the DiT predicts the noise residual conditioned on both semantic and visual features. After denoising, the recovered latent $\bm{x}_0$ is passed through the VAE decoder to obtain the final edited image $I^{\text{edit}}$.

\tit{Optimization Objective}
The model is optimized via the standard diffusion loss $\mathcal{L}_{\mathrm{diffusion}} =  \mathbb{E}_{\bm{x}_0, \bm{x}_t, t} \left[\left\|\bm{x}_t - \hat{\bm{x}}_t\right\|^2\right]$.

\section{Experiments}
\label{sec:experiments}
\vspace{-7pt}
We conduct extensive experiments on \dataset to validate our proposed approach and to establish benchmark results for instruction-driven fashion editing. All evaluations follow the benchmark protocols introduced in Sec.~\ref{subsec:benchmarking}. Results are reported for instruction-driven VTON in both paired and unpaired configurations, as well as instruction-driven VTOFF. We use the real images as the ground truth to ensure a proper evaluation that is not influenced by the quality upper bounds of the ground-truth synthetic images. Therefore, given a synthetic image $I^{\text{edit}}_{\text{person}}$, we use the instruction $T^{r}_{\text{inst}}$ to edit the image in order to predict $I_{\text{person}}$.
\vspace{-7pt}
\subsection{Implementation and Training Details}
\label{subsec:training_details}
\vspace{-5pt}
All models are trained on the proposed dataset using a resolution of 768$\times$1024. We employ AdamW~\cite{loshchilov2019decoupledweightdecayregularization} as optimizer, using a cosine scheduler with a learning rate of 1~$\times$10$^{-4}$ and a batch size of 4 for 50k steps. We use DeepSpeed~\cite{rajbhandari2020zeromemoryoptimizationstraining} and train on 16 NVIDIA A100 GPUs. As diffusion backbone, we start from Stable Diffusion 3 medium~\cite{esser2024scaling}, while we employ InternVL-3.5-8B~\cite{wang2025internvl3} as MLLM. During training, the VAE and the MLLM are kept frozen, while the connector and the DiT are trained end-to-end. During training, we utilize both original and edited images as ground-truth targets. This mixed supervision strategy forces the model to maintain robust alignment with real-world textures while simultaneously learning complex, instruction-driven structural and appearance modifications.

\subsection{Baselines and Competitors}
We compare our method against both task-specific and general-purpose diffusion-based approaches relevant to VTON, VTOFF, and instruction-guided editing. For fairness, all task-specific models are retrained on \dataset using either their official source codes, when available, or faithful re-implementations following the original architectures and training protocols. For all comparisons, input resolution and evaluation splits are kept identical across methods.

\begin{table}[t]
    \centering
    \setlength{\tabcolsep}{.38em}
    \caption{Quantitative results on instruction-driven VTON (paired, unpaired) and VTOFF on \dataset. We report FID, KID, SSIM, LPIPS, DISTS, and DINO-I. $\downarrow$ indicates lower is better, $\uparrow$ indicates higher is better.}
    \vspace{-0.15cm}
    \resizebox{\linewidth}{!}{
    \begin{tabular}{lc ccc cc cc cc c}
    \toprule
    & & \multicolumn{6}{c}{\textbf{Paired}} & & \multicolumn{3}{c}{\textbf{Unpaired}} \\
    \cmidrule{3-8} \cmidrule{10-12}
    \textbf{Method} & & SSIM $\uparrow$ & LPIPS $\downarrow$ & DISTS $\downarrow$ & FID $\downarrow$ & KID $\downarrow$ & DINO-I $\uparrow$ & & FID $\downarrow$ & KID $\downarrow$ & DINO-I $\uparrow$\\
    \midrule
    \rowcolor{cyan!5}
    \multicolumn{12}{c}{\textbf{Edited VTON}} \\
    \midrule
        FLUX.2 Klein~\cite{flux-2-2025} & & 0.9291 & 0.0743 & 0.0853 & 4.32 & 1.71 & 0.9302 & & 6.29 & 1.80 & \textbf{0.6549}\\
        Qwen-Image-Edit~\cite{wu2025qwen} & & 0.9173 & 0.0963 & 0.1135 &  8.51 & 3.42 & 0.8802 & & 10.19 & 3.24 & 0.6453 \\    
        Any2AnyTryon~\cite{guo2025any2anytryonleveragingadaptiveposition} & & \textbf{0.9429} & 0.0702 & 0.0795 & 5.34 & 1.98 & 0.9074 & & 6.36 & 1.94 & 0.4996\\
        CatVTON~\cite{chong2025catvton} & & 0.9314 & 0.0801 & 0.1022 & 4.57 & 1.72 & 0.9144 & & 6.40 & 1.91 & 0.6476\\
    \rowcolor{cyan!15}
        \textbf{\ours (Ours)} & & 0.9417 & \textbf{0.0628} & \textbf{0.0733} & \textbf{3.79} & \textbf{1.13} & \textbf{0.9383} & & \textbf{5.82} & \textbf{1.44} & 0.6476\\
    \addlinespace[2.5pt]
    \midrule
    \addlinespace[2.5pt]
    \rowcolor{teal!5}
    \multicolumn{12}{c}{\textbf{Edited VTOFF}} \\
    \midrule
        FLUX.2 Klein~\cite{flux-2-2025} & & 0.8694 & 0.2800 & 0.2579 & 17.45 & 7.48 & 0.7468 & & - & - & -\\
        Qwen-Image-Edit~\cite{wu2025qwen} & & 0.8463 & 0.2970 & 0.2792 &  24.09 & 9.56 & 0.7192 & & - & - & - \\
        Any2AnyTryon~\cite{guo2025any2anytryonleveragingadaptiveposition} & & 0.8757 & 0.2878 & 0.2474 & 19.03 & 6.26 & 0.6864 & & - & - & -\\
        CatVTON~\cite{chong2025catvton} & & \textbf{0.8800} & 0.2415 & 0.2058 & 6.78 & 1.45 & 0.7918 & & - & - & -\\
    \rowcolor{teal!15}
        \textbf{\ours (Ours)} & & \textbf{0.8800} & \textbf{0.2060} & \textbf{0.1892} & \textbf{5.06} & \textbf{0.73} & \textbf{0.8071} & & - & - & -\\
    \bottomrule
    \end{tabular}
    }
    \label{tab:experiments_results_reverse}
    \vspace{-0.3cm}
\end{table}

\tit{VTON and VTOFF Models}
We include recent state-of-the-art approaches for standard VTON and VTOFF, namely CatVTON~\cite{chong2025catvton} and Any2AnyTryon~\cite{guo2025any2anytryonleveragingadaptiveposition}.  Since CatVTON does not natively support text input, we extend its architecture by incorporating instruction embeddings following standard conditioning practices in diffusion-based editing. This adaptation enables fair comparison under instruction-driven settings. To ensure consistency, CatVTON is retrained with Stable Diffusion~3 (medium variant), matching the pretrained backbone used in our \ours model.

\tit{General-purpose Editing Models}
We further evaluate large-scale multimodal diffusion models in a zero-shot setting to examine their capability for instruction-driven garment manipulation. Specifically, we include FLUX.2 Klein~\cite{flux-2-2025} and Qwen-Image-Edit~\cite{wu2025qwen}, both capable of text-guided editing across generic domains. These models serve as strong foundation-level baselines for assessing semantic alignment and edit fidelity in complex fashion edits.
A critical challenge is to extend these models to multi-image scenarios such as VTON. We address this scenario by concatenating image tokens at the input of the DiT~\cite{peebles2023scalable} for both models, while preserving their spatial relationships through positional encoding.

\begin{table*}[t]
    \centering
    \setlength{\tabcolsep}{.2em}
    \caption{Per-edit-type quantitative results for instruction-driven VTON and VTOFF on \dataset. We report DISTS ($\downarrow$) and DINO-I ($\uparrow$) for the categories Change Color, Change Pattern, Change Material, Fine-Grained, Add Detail, Remove Element, and Modify Structure. 
    }
    \vspace{-0.15cm}
    \resizebox{\linewidth}{!}{
    \begin{tabular}{lc cc c cc c cc c cc c cc c cc c cc}
    \toprule
    & & \multicolumn{2}{c}{\textbf{Color}} & & \multicolumn{2}{c}{\textbf{Pattern}} & & \multicolumn{2}{c}{\textbf{Material}} & & \multicolumn{2}{c}{\textbf{Fine-Gr.}} & & \multicolumn{2}{c}{\textbf{Add}} & & \multicolumn{2}{c}{\textbf{Remove}} & & \multicolumn{2}{c}{\textbf{Struct}} \\
    \cmidrule{3-4} \cmidrule{6-7} \cmidrule{9-10} \cmidrule{12-13} \cmidrule{15-16} \cmidrule{18-19} \cmidrule{21-22}
    \textbf{Method} & & DISTS & DINO & & DISTS & DINO & & DISTS & DINO & & DISTS & DINO & & DISTS & DINO & & DISTS & DINO & & DISTS & DINO \\
    \midrule
    \rowcolor{cyan!5}
    \multicolumn{22}{c}{\textbf{Edited VTON}} \\
    \midrule
        FLUX.2 Klein~\cite{flux-2-2025} & & 0.081 & 0.939 & & 0.100 & 0.908 & & 0.085 & 0.939 & & 0.081 & 0.936 & & 0.075 & 0.941 & & 0.084 & 0.935 & & 0.078 & 0.943 \\
        Qwen-Image-Ed~\cite{wu2025qwen} & & 0.106 & 0.896 & & 0.171 & 0.820 & & 0.093 & 0.893 & & 0.089 & 0.922 & & 0.086 & 0.894 & & 0.082 & 0.926 & & 0.080 & 0.941  \\
        Any2AnyTryon~\cite{guo2025any2anytryonleveragingadaptiveposition} & & 0.070 & 0.937 & & 0.099 & 0.865 & & 0.072 & 0.923 & & 0.076 & 0.917 & & 0.074 & 0.906 & & \textbf{0.068} & 0.940 & & 0.074 & 0.929  \\
        CatVTON~\cite{chong2025catvton} & & 0.096 & 0.929 & & 0.117 & 0.889 & & 0.099 & 0.922 & & 0.098 & 0.912 & & 0.096 & 0.919 & & 0.098 & 0.931 & & 0.097 & 0.921 \\
    \rowcolor{cyan!15}
        \textbf{\ours (Ours)} & & \textbf{0.066} & \textbf{0.949} & & \textbf{0.087} & \textbf{0.920} & & \textbf{0.070} & \textbf{0.942} & & \textbf{0.072} & \textbf{0.938} & & \textbf{0.067} & \textbf{0.944} & & 0.070 & \textbf{0.947} & & \textbf{0.069} & \textbf{0.946} \\

    \addlinespace[2.5pt]
    \midrule
    \addlinespace[2.5pt]
    \rowcolor{teal!5}
    \multicolumn{22}{c}{\textbf{Edited VTOFF}} \\
    \midrule
        FLUX.2 Klein~\cite{flux-2-2025} & & 0.252 & 0.757 & & 0.299 & 0.707 & & 0.245 & 0.755 & & 0.227 & 0.790 & & 0.237 & 0.754 & & 0.255 & 0.728 & & 0.232 & 0.816 \\
        Qwen-Image-Ed~\cite{wu2025qwen} & & 0.267 & 0.757 & & 0.379 & 0.650 & & 0.238 & 0.757 & & 0.233 & 0.761 & & 0.227 & 0.725 & & 0.260 & 0.729 & & 0.221 & 0.766 \\
        Any2AnyTryon~\cite{guo2025any2anytryonleveragingadaptiveposition} & & 0.226 & 0.765 & & 0.267 & 0.699 & & 0.220 & 0.769 & & 0.228 & 0.775 & & 0.238 & 0.707 & & 0.232 & 0.745 & & 0.227 & 0.762 \\
        CatVTON~\cite{chong2025catvton} & & 0.210 & 0.807 & & 0.236 & 0.747 & & 0.197 & 0.817 & & 0.197 & 0.802 & & 0.181 & 0.807 & & 0.204 & 0.804 & & 0.187 & 0.808 \\
    \rowcolor{teal!15}
        \textbf{\ours (Ours)} & & \textbf{0.180} & \textbf{0.828} & & \textbf{0.219} & \textbf{0.771} & & \textbf{0.183} & \textbf{0.827} & & \textbf{0.181} & \textbf{0.821} & & \textbf{0.174} & \textbf{0.811} & & \textbf{0.187} & \textbf{0.828} & & \textbf{0.172} & \textbf{0.819}  \\
    \bottomrule
    \end{tabular}
    }
    \label{tab:results_edit_types}
    \vspace{-0.3cm}
\end{table*}

\begin{figure*}[t]
    \centering
    \includegraphics[width=0.98\linewidth]{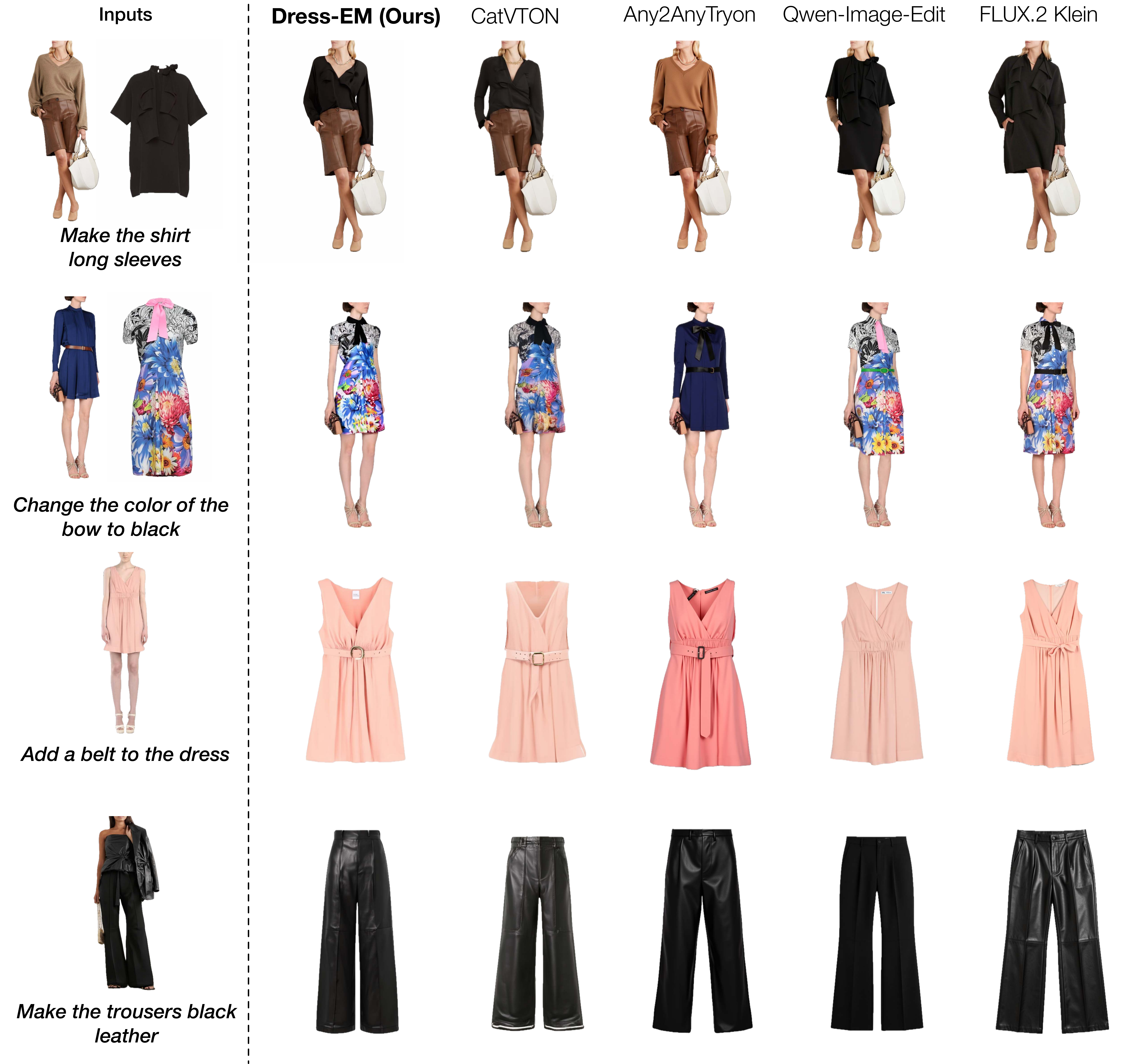}
    \vspace{-0.2cm}
    \caption{Qualitative results for edited VTON unpaired setting (first two rows) and edited VTOFF (last two rows), showing realistic and instruction-consistent edits.}
    \label{fig:qualitative_edits}
    \vspace{-0.3cm}
\end{figure*}

\subsection{Editing Results}

We report quantitative and qualitative comparisons between \ours and competing methods across all benchmark settings, using the metrics defined in Sec.~\ref{subsec:benchmarking}.  

\tit{Quantitative Results}
Table~\ref{tab:experiments_results_reverse} summarizes quantitative results across all editing tasks and settings. Overall, \ours consistently outperforms all other methods, demonstrating strong generalization across both paired and unpaired scenarios.
In the paired edited VTON setting, \ours achieves the best results across nearly all metrics, including the lowest FID, KID, LPIPS, and DISTS values. These results indicate superior image realism and faithful preservation of garment structure and identity compared to CatVTON~\cite{chong2025catvton} and Any2AnyTryon~\cite{guo2025any2anytryonleveragingadaptiveposition}.  
The strong DINO-I further validates the ability of the model to generate precise and semantically consistent edits.
In the more challenging unpaired edited VTON configuration, specialized models demonstrate robust performance, maintaining a competitive edge even when handling garments unseen during the original try-on process. \ours maintains high DINO-I value despite the increased complexity, confirming the robustness of its MLLM-guided conditioning and connector design.

For the edited VTOFF task, which reconstructs the edited in-shop garment from the dressed person, \ours confirms the best overall performance across nearly all metrics, including the lowest FID, KID, and DISTS, and the highest DINO-I. These results demonstrate its strong capability to recover fine garment details and produce semantically faithful edits.

\tit{Per-Edit-Type Analysis}
Table~\ref{tab:results_edit_types} reports a fine-grained breakdown across the seven edit categories defined in \dataset, evaluated using DISTS and DINO-I metrics. \ours consistently achieves the lowest DISTS values (indicating higher perceptual similarity) and the highest DINO-I scores across nearly all edit types for both VTON and VTOFF. Notably, the largest improvements occur in appearance-driven edits such as color, pattern, and material, where multimodal conditioning effectively captures localized texture and tone changes.  Structural modifications, including sleeve length and shape adjustments, show consistent quantitative gains, indicating that our model effectively preserves spatial coherence while performing localized structural edits.

\tit{Qualitative Results} A visual comparison is presented in Fig.~\ref{fig:qualitative_edits}, which showcases representative examples across both VTON (unpaired) and VTOFF settings. As it can be seen, \ours produces high-quality and photorealistic results that accurately follow the editing instructions while preserving person identity, pose, and garment details, further confirming the effectiveness of the proposed approach.
\vspace{-7pt}

\section{Conclusion}
\label{sec:conclusion}
\vspace{-8pt}
We introduced \dataset, the first large-scale dataset and benchmark for instruction-driven fashion editing that unifies VTON, VTOFF, and text-guided garment modification within a single framework. Built through a fully automated multimodal pipeline integrating MLLM-based garment understanding, diffusion-based synthesis, and LLM-guided verification, \dataset comprises over 146k verified, semantically consistent samples across diverse edit types. Beyond dataset construction, it establishes unified benchmarks and a multimodal diffusion baseline, enabling systematic evaluation of controllable and interpretable fashion generation. We hope this work fosters future research on multimodal garment understanding and interactive editing.

\tit{Acknowledgments}
This work was supported by EU Horizon projects ELIAS (No. 101120237) and ELLIOT (No. 101214398), and by the FIS project GUIDANCE (No. FIS2023-03251). We also acknowledge the CINECA award under the ISCRA initiative, for the availability of high-performance computing resources.

\bibliographystyle{splncs04}
\bibliography{bibliography}

\clearpage
\appendix

\section{Additional Implementation Details}
\label{supp:details}

\tinytit{Training Configuration}
During training, we employ DeepSpeed~\cite{rajbhandari2020zeromemoryoptimizationstraining} for memory optimization and use \texttt{bfloat16} precision to ensure training stability and efficiency. %
As mentioned in the main paper, the weights of the VAE and the MLLM are kept frozen. Only the connector module and the DiT parameters are optimized end-to-end.

\section{Additional Dataset Details}
\label{supp:dataset_details}

\subsection{Mask Preprocessing}
FitDiT uses bounding boxes as input masks. These bounding boxes are extracted from the label maps of a clothed person extracted using a parser. A visual 
\begin{wrapfigure}{r}{0.52\linewidth}
    \centering
    \vspace{-12pt}
    \includegraphics[width=0.98\linewidth]{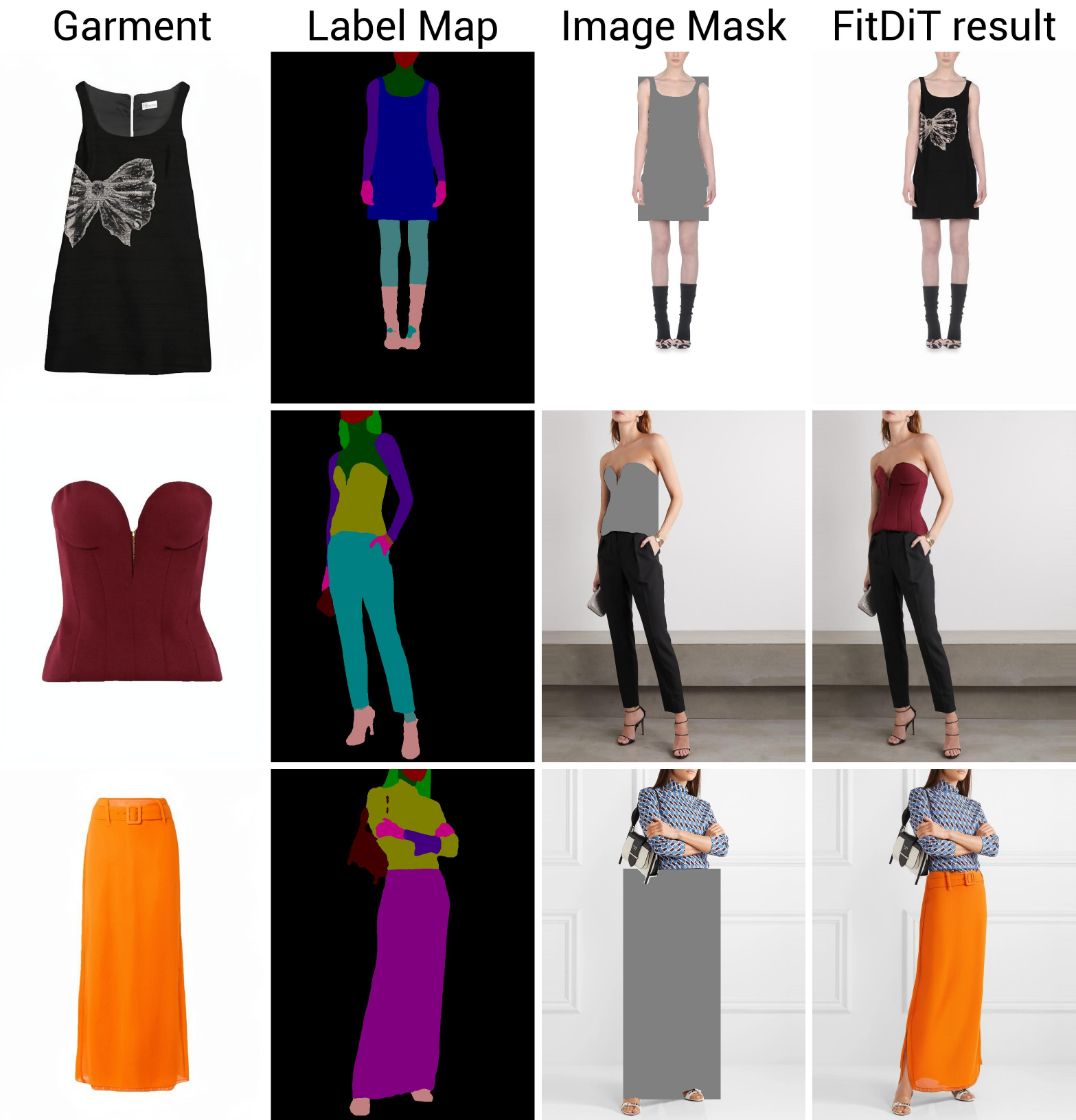}
     \vspace{-3pt}
    \caption{Sample of label maps, masks, in-shop garment and FitDiT result.}
    \label{fig:mask_fitdit}
     \vspace{-15pt}
\end{wrapfigure}
example is provided in Fig.~\ref{fig:mask_fitdit}
The label maps provided in Dress Code~\cite{morelli2022dresscode} have significant inconsistencies in the hand area, partially masking hands too. More powerful segmentation networks such as Sapiens~\cite{khirodkar2024sapiens} are better at segmenting body parts but struggle to isolate garments from each other, resulting in artifacts, for example upper-body and lower-body items labeled as one.
For this reason, we employ a new segmentation model~\cite{fashn-human-parser}, specifically trained on garment data and capable of isolating body parts. We then extract the bounding box mask from the new segmentation maps and apply FitDiT~\cite{jiang2024fitdit} to virtually try-on the edited ground-truth garment to the person, while preserving hands.

\subsection{Rule-based Templates}
The dataset generation pipeline employs Qwen3-VL~\cite{bai2025qwen3} to extract garment attributes, which are converted into paired editing instructions via rule-based templates ($T_{inst}$).  
For each valid edit, we generate both a forward and a reverse instruction (to undo or invert the transformation), with category-aware constraints for upper-body, lower-body, and dresses. Our final dataset includes seven distinct edit types.

\tit{Change Color}
This edit is generated only for solid garments. A target color is sampled from a predefined palette, excluding colors already present in the source description:
``\textit{Change the color of the garment to} \texttt{[Target Color]}''.

\tit{Change Pattern}
A new pattern is sampled from a predefined pattern set (excluding the source pattern). The instruction depends on the source pattern:
if the source is solid, the edit is additive:
``\textit{Add a} \texttt{[Target Pattern]} \textit{pattern to the garment}'';  
otherwise, it is a replacement:
``\textit{Replace the} \texttt{[Source Pattern]} \textit{pattern with a} \texttt{[Target Pattern]} \textit{pattern}''.

\tit{Change Material}
This edit is also restricted to solid garments. The target material is sampled from a material pool excluding the source, while preserving garment color in the prompt:
``\textit{Make the garment} \texttt{[Garment Color]} \texttt{[Target Material]}''.

\tit{Modify Structure}
A structural attribute (\eg, sleeves or neckline) is selected from category-specific valid attributes, and a different target value is sampled:
``\textit{Change the} \texttt{[Attribute]} \textit{to} \texttt{[Target Value]}''.

\tit{Add Detail}
A new distinctive feature is sampled from category-specific addable features, excluding features already present on the garment.  
Generic form:
``\textit{Add a} \texttt{[Feature]} \textit{to the garment}'',  
or, when location is defined:
``\textit{Add a} \texttt{[Feature]} \textit{at the} \texttt{[Location]} \textit{of the garment}''.

\tit{Remove Element}
A removable feature is selected from \textit{verified} distinctive features that satisfy category and location-specific constraints:
``\textit{Remove the} \texttt{[Feature]} \textit{from the garment}''.

\tit{Fine-Grained} It consists of coloring small details or feature replacement.
For \textit{Color Detail}, a color-editable feature is selected from \textit{verified} distinctive features under category/location constraints, and recolored to a new color:
``\textit{Change the color of the} \texttt{[Feature]} \textit{to} \texttt{[Target Color]}''. Instead, for \textit{Replace Detail}, a feature replacement follows predefined category-specific\\ source$\rightarrow$target pairs (\eg, bow$\rightarrow$buttons, belt$\rightarrow$drawstring), used only when the required source feature (and location, if needed) exists:
``\textit{Replace the} \texttt{[Source Feature]} \textit{with} \texttt{[Target Feature]}''.

\begin{figure*}[t]
    \centering
    \includegraphics[width=\linewidth]{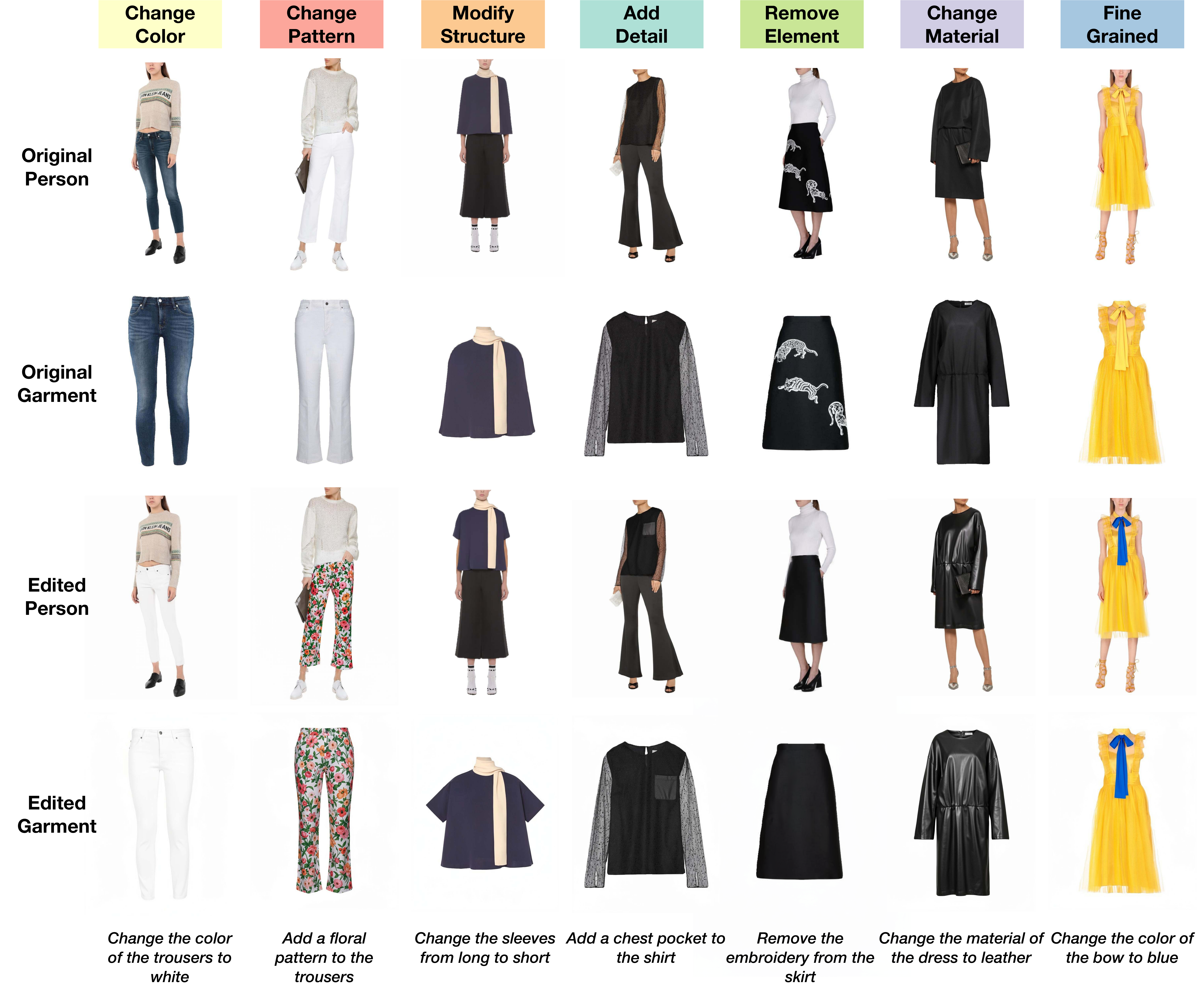}
    \vspace{-0.5cm}
    \caption{Samples from \dataset across the different edit types, including the original person-garment pair, the corresponding edited outputs, and the textual instruction.}
    \vspace{-0.3cm}
 \label{fig:supp_dataset_visual_examples}
\end{figure*}

\subsection{Attribute Selection} 
To ensure diversity while preserving plausibility, template slots are populated from predefined attribute banks with category-aware constraints:
\smallbreak\noindent\textbf{Colors (global and detail-level):} \textit{red}, \textit{blue}, \textit{green}, \textit{yellow}, \textit{black}, \textit{white}, \textit{purple}, \textit{orange}, \textit{pink}, \textit{brown}, \textit{gray}, \textit{navy}, \textit{beige}, \textit{lavender}, \textit{maroon}, \textit{teal}.
\smallbreak\noindent\textbf{Patterns:} \textit{polka dot}, \textit{striped}, \textit{floral}, \textit{geometric}, \textit{checkered}, \textit{paisley}, \textit{camouflage}, \textit{leopard print}, \textit{gingham}, \textit{houndstooth}, \textit{abstract}.
\smallbreak\noindent\textbf{Materials:} \textit{denim}, \textit{leather}, \textit{silk}, \textit{wool}, \textit{cotton}, \textit{linen}.
\smallbreak\noindent\textbf{Structural Attributes:} \textit{long sleeves}, \textit{short sleeves}, \textit{sleeveless}, \textit{cap sleeves}, \textit{puffy sleeves}, \textit{three-quarter sleeves}, \textit{v-neck}, \textit{crew neck}, \textit{scoop neck}, \textit{square neck}, \textit{high neck}, \textit{collar}.
\smallbreak\noindent\textbf{Addable Details (category-specific)}
\begin{itemize}[noitemsep, topsep=0pt]
    \item \textit{Upper-body:} \textit{bow} (neck), \textit{pockets}, \textit{chest pocket}, \textit{zipper}, \textit{buttons}.
    \item \textit{Lower-body:} \textit{belt}, \textit{drawstring}.
    \item \textit{Dresses:} \textit{belt}, \textit{fitted belt}, \textit{pockets}, \textit{drawstring}, \textit{bow} (neck/waist).
\end{itemize}
\smallbreak\noindent\textbf{Removable/Colorable/Replaceable Details:} sampled from verified distinctive features under category- and location-specific rules (\eg, \textit{logo}, \textit{text}, \textit{zipper}, \textit{drawstring}, \textit{pocket}, \textit{embroidery}, \textit{fringe}, \textit{belt}, \textit{bow}, \textit{buttons}, \textit{sash}, \textit{stripes}, \textit{feather trim}), including predefined replacement pairs (\eg, \textit{belt} $\leftrightarrow$ \textit{bow}, \textit{belt} $\leftrightarrow$ \textit{drawstring}, \textit{bow} $\leftrightarrow$ \textit{buttons}).

\begin{wraptable}{r}{0.32\linewidth}
    \centering
    \vspace{-1.1cm}
    \resizebox{0.95\linewidth}{!}{
    \begin{tabular}{lrr}
    \toprule
    \textbf{Category} & \textbf{Count} & \textbf{(\%)} \\
    \midrule
    Dresses & 80,865 & 55.2 \\
    Upper-body & 45,567 & 31.1 \\
    Lower-body & 20,028 & 13.7 \\
    \midrule
    \textbf{Total} & 146,460 & 100.0 \\
    \bottomrule
    \end{tabular}}
    \vspace{-0.15cm}
    \caption{Sample distribution by garment category.}
    \label{tab:category_stats}
    \vspace{-0.5cm}
\end{wraptable}

\subsection{Dataset Examples and Statistics}
\begin{figure}[t]
    \centering
    \includegraphics[width=0.8\linewidth]{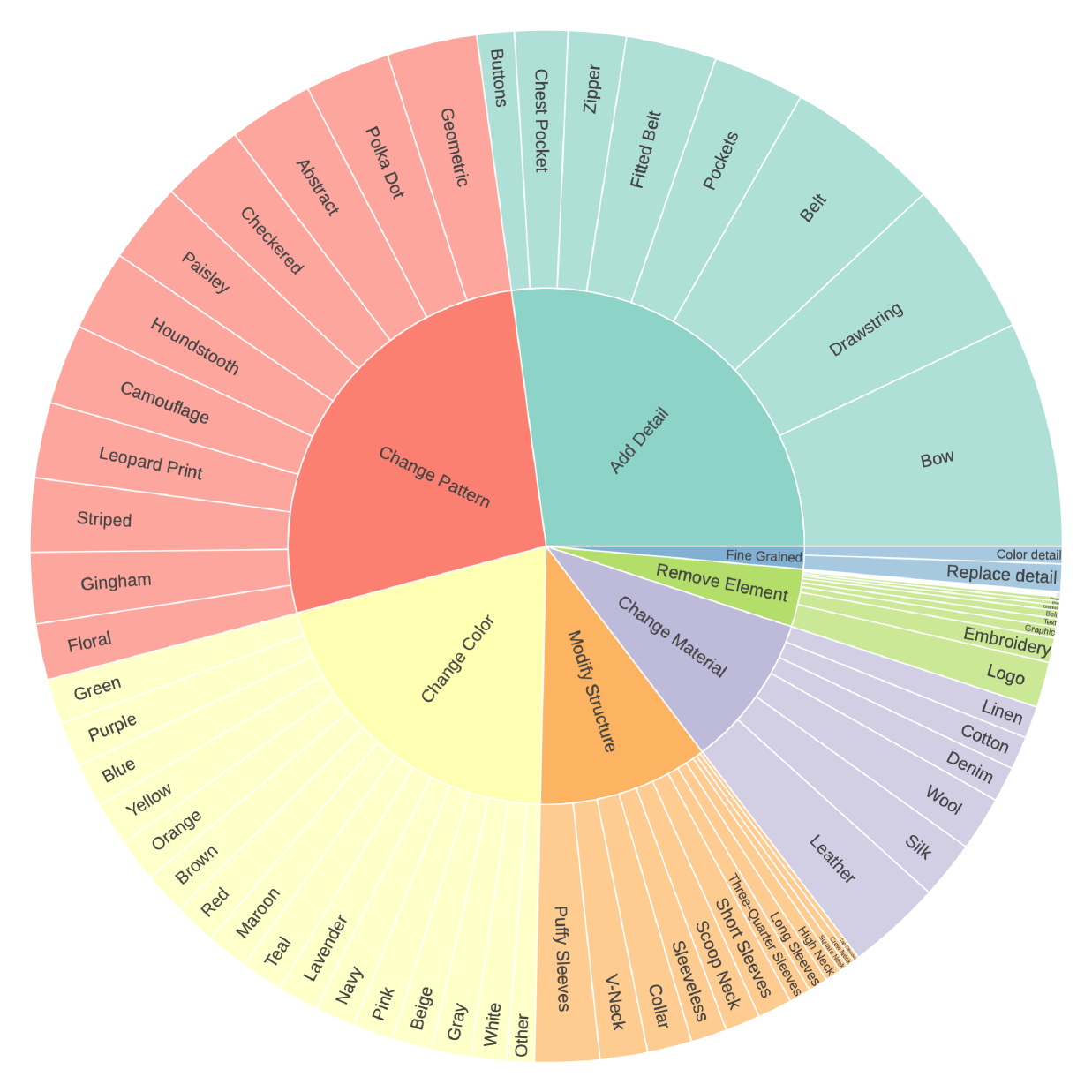}
    \vspace{-0.3cm}
    \caption{Editing instruction taxonomy showing the distribution of edit types in \dataset, covering both appearance-level changes (\eg, color, pattern, material) and structural modifications (\eg, sleeve length, neckline, shape).}
    \label{fig:edit_taxonomy}
    \vspace{-0.35cm}
\end{figure}

To provide further details on \dataset, we report the visual taxonomy of our data in Fig.~\ref{fig:edit_taxonomy}.
As mentioned in the main paper, the dataset follows the standard split of Dress Code~\cite{morelli2022dresscode}, ensuring no overlap of garment identities between training and testing sets. Table~\ref{tab:category_stats} reports the distribution of samples across garment categories.
Additional examples from \dataset are provided in Fig.~\ref{fig:supp_dataset_visual_examples}, showing the original person-garment pair, the corresponding instruction, and the resulting edited outputs.

\section{Additional Quantitative Results}
\tinytit{Forward vs. Inverse Editing}
To address artifact inversion and validate the generalization of our approach, we compare forward editing (real $\rightarrow$ synthetic) and inverse editing (synthetic $\rightarrow$ real, as in the main protocol). Results are shown in Table~\ref{tab:experiments_results_supp}, considering both edited VTON and VTOFF on the paired test set of \dataset. As it can be seen, this comparison shows almost identical results across directions, indicating that the model learns semantic editing rather than simply reversing artifacts. This is further supported by our mixed supervision strategy, which includes real images as targets, preventing collapse to the distribution of synthetic images.

\tit{Comparison with Commercial Models}
To assess the performance of commercial models, we compare \ours with GPT-Image-1.5 on a random 100-sample subset of the test set (paired edited VTON). Results, reported in Table~\ref{tab:experiments_gpt}, show that, despite their generality, frontier image editing models underperform compared to a domain-specific model trained on \dataset, highlighting the relevance of the proposed benchmark.

\begin{table}[t]
%\vspace{-0.4cm}
    \centering
    \setlength{\tabcolsep}{.4em}
    \caption{Forward vs. reverse editing on \dataset (paired).}
    \vspace{-0.15cm}
    \resizebox{\linewidth}{!}{
    \begin{tabular}{lc cccc c cccc}
    \toprule
    & \multicolumn{4}{c}{\cellcolor{cyan!5}\textbf{Edited VTON}} & & \multicolumn{4}{c}{\cellcolor{teal!5}\textbf{Edited VTOFF}} \\
    \cmidrule{2-5} \cmidrule{7-10}
    & SSIM $\uparrow$ & LPIPS $\downarrow$ & DISTS $\downarrow$ & DINO-I $\uparrow$ & & SSIM $\uparrow$ & LPIPS $\downarrow$ & DISTS $\downarrow$ & DINO-I $\uparrow$ \\
    \midrule
    Forward & 0.9317 & 0.0627 & 0.0724 & 0.9493 & & 0.8454 & 0.2084 & 0.1871 & 0.8212 \\
    Inverse & 0.9417 & 0.0628 & 0.0733 & 0.9383 & & 0.8800 & 0.2060 & 0.1892 & 0.8071 \\
    \bottomrule
    \end{tabular}
    }
    \label{tab:experiments_results_supp}
    \vspace{-0.15cm}
\end{table}

\begin{table}[t]
    \centering
    \setlength{\tabcolsep}{.4em}
    \caption{Comparison with commercial models on edited VTON.}
    \vspace{-0.15cm}
    \resizebox{0.66\linewidth}{!}{
    \begin{tabular}{l cccc}
    \toprule
     & SSIM $\uparrow$ & LPIPS $\downarrow$ & DISTS $\downarrow$ & DINO-I $\uparrow$ \\
    \midrule
    \textbf{\ours (Ours) }  & \textbf{0.9437} & \textbf{0.0622} & \textbf{0.0721} & \textbf{0.9415} \\
    GPT-Image-1.5  & 0.9323 & 0.0838 & 0.0842 & 0.9406 \\
    \bottomrule
    \end{tabular}
    }
    \label{tab:experiments_gpt}
    \vspace{-0.3cm}
\end{table}

\section{Additional Qualitative Results}
\label{supp:qualitatives}
\begin{figure}[t]
    \centering
    \includegraphics[width=\linewidth]{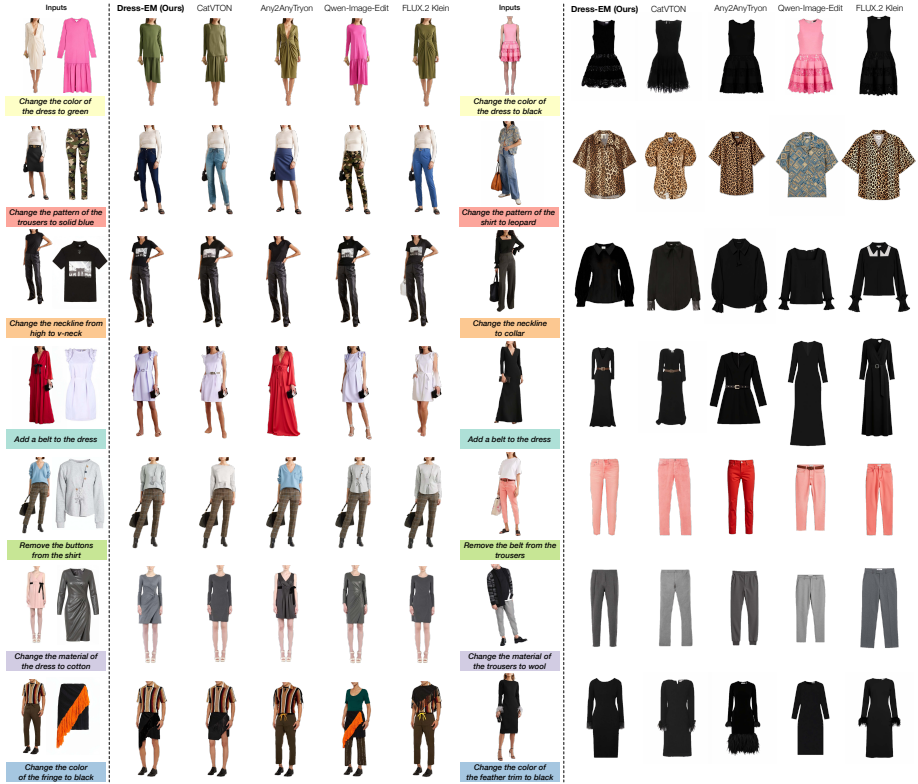}
    \vspace{-0.5cm}
    \caption{Qualitative results for edited VTON (left, unpaired setting) and VTOFF (right), showing realistic and instruction-consistent edits.}
    \label{fig:supp_qualitative_edits}
    \vspace{-0.3cm}
\end{figure}

In Fig.~\ref{fig:supp_qualitative_edits}, we provide additional qualitative comparisons for instruction-driven VTON and VTOFF across the range of edit types included in \dataset. We compare our \ours model with task-specific methods retrained on \dataset\ (\ie, CatVTON~\cite{chong2025catvton} and Any2AnyTryon~\cite{guo2025any2anytryonleveragingadaptiveposition}) as well as general-purpose editing models evaluated in a zero-shot setting (Qwen-Image-Edit~\cite{wu2025qwen} and FLUX.2~Klein~\cite{flux-2-2025}). Zero-shot models typically struggle to preserve garment identity, maintain viewpoint consistency, or follow the instruction faithfully. Retrained VTON/VTOFF baselines provide more coherent edits, but still exhibit artifacts or incomplete transformations. In contrast, \ours produces visually realistic and instruction-consistent results across appearance and structural edits, confirming the trends observed in the quantitative and qualitative analyses reported in the main paper.  
    
\section{Additional Discussion and Clarifications}
\label{supp:more_discuss}
This section provides further explanations on several components of our pipeline. We outline the motivations behind key design choices, describe their practical implications, and discuss limitations relevant to interpreting the dataset and evaluation results.

\subsection{On the Use of Synthetic Edited Images}
The edited garments and try-on results in \dataset are produced through a controlled multimodal generation pipeline. This choice is deliberate: instruction-driven fashion editing requires tightly aligned pairs of garment, person, instruction, and edited outputs, which are extremely difficult to obtain consistently through manual annotation. Generative editing ensures that the modification directly corresponds to the instruction and that the garment identity is preserved before and after editing, which is essential for evaluating semantic edit correctness.

Rather than replacing in-the-wild data, the synthetic construction complements it. \dataset\ provides a controlled and reproducible benchmark for measuring whether models follow the intended transformation, while remaining compatible with future real-world data collection or domain adaptation. In this sense, the dataset serves as a structured foundation on top of which broader and more diverse fashion-editing scenarios can be explored.

\begin{figure}[t]
    \centering
    \includegraphics[width=\linewidth]{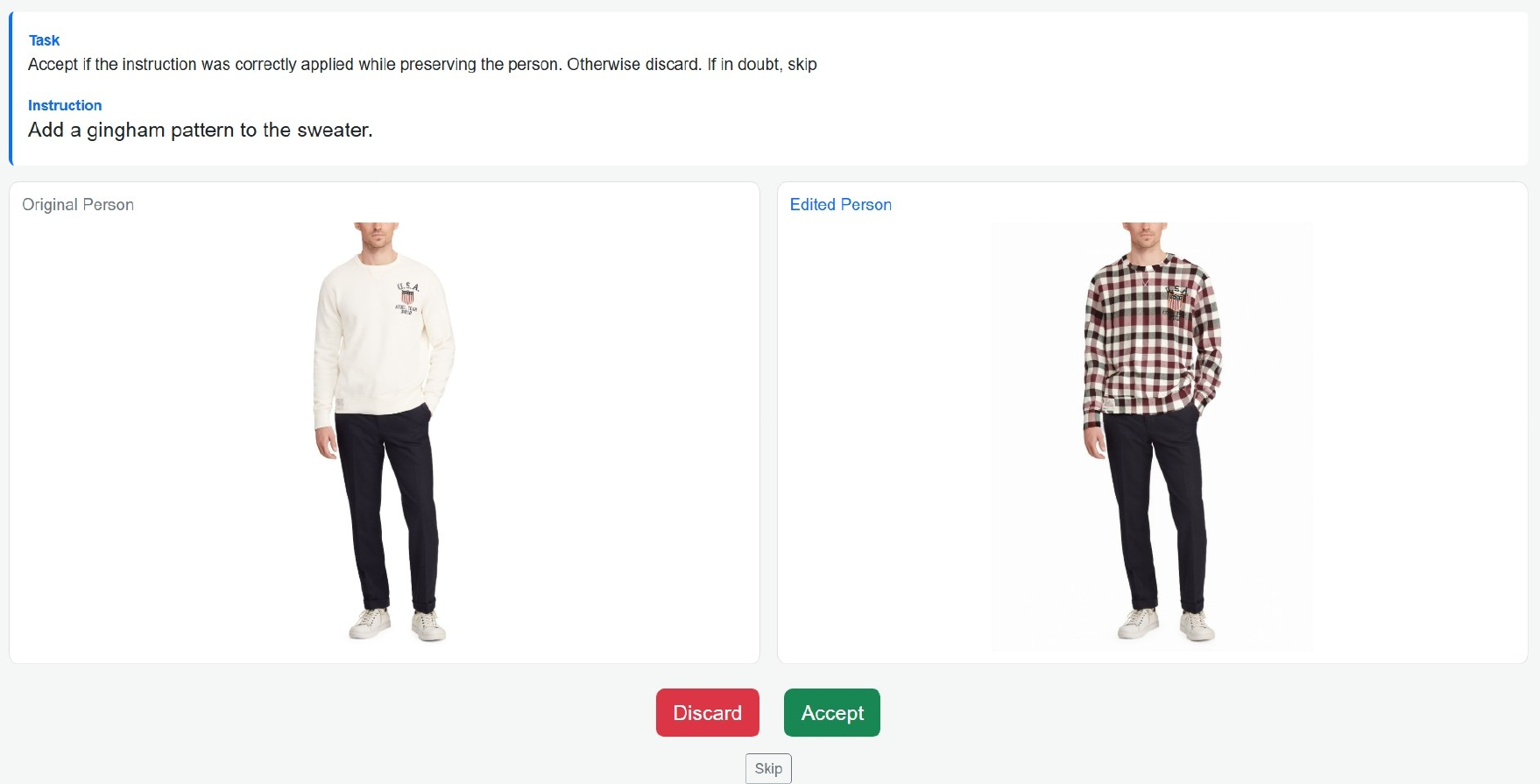}
    \vspace{-0.5cm}
    \caption{User study interface.}
    \label{fig:user_study}
    \vspace{-0.35cm}
\end{figure}

\subsection{LLM-as-Judge and Evaluation Consistency}
GPT-5~\cite{openaigpt-5} is employed solely as a \emph{non-generative evaluator}: it does not participate in any image synthesis, nor does it influence VTON/VTOFF models. Its role is restricted to assessing whether an edited result follows the instruction while preserving garment identity and visual plausibility. Using a strong evaluator at this stage helps establish clean supervision signals for filtering, especially for fine-grained structural edits that are difficult to evaluate automatically.

To enable reproducible large-scale verification, we distill GPT-5's judgments into a smaller open-source MLLM (InternVL-3.5~\cite{wang2025internvl3}). Although this introduces some dependence between data construction and evaluation, our final benchmark reports model-agnostic evaluation metrics (\ie, including DINO and standard perceptual scores) which remain independent of both GPT-5 and InternVL. This design ensures that LLM-based judging improves data reliability without dominating or biasing downstream evaluation.

\tit{User Study on MLLM Filtering Scores}
To evaluate the coherence of our fine-tuned MLLM, we conducted a user study using the interface illustrated in Fig.~\ref{fig:user_study}. Participants were tasked with assessing the edit correctness of triplets 
\begin{wrapfigure}[14]{r}{0.5\linewidth}
    \centering
    \vspace{-0.7cm}
    \includegraphics[width=0.98\linewidth]{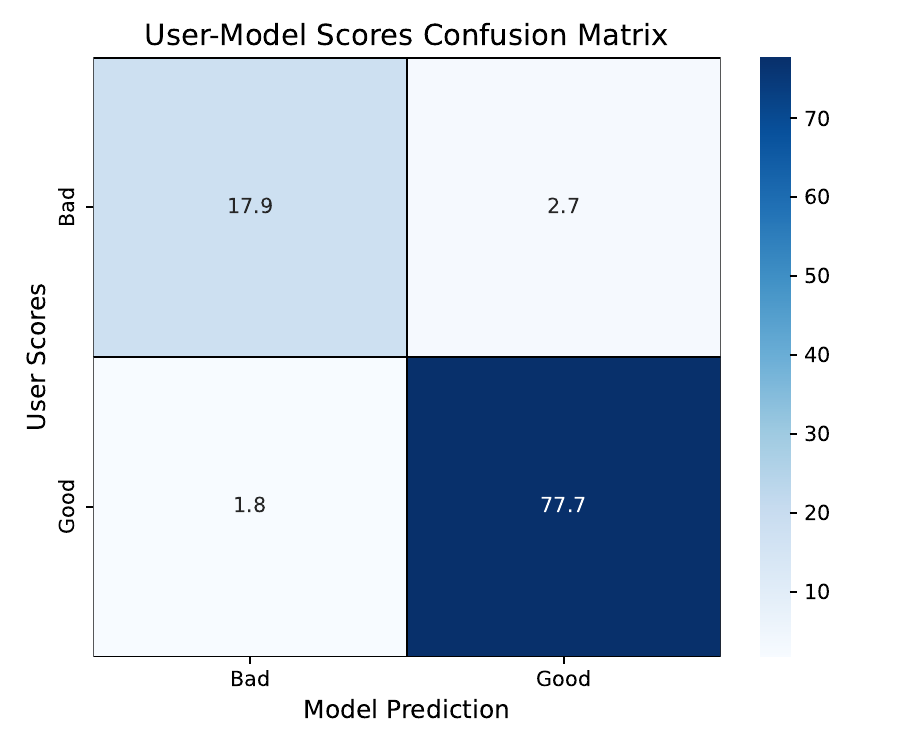}
    \vspace{-0.3cm}
    \caption{Confusion matrix of our user study.}
    \label{fig:confusion_matrix}
\end{wrapfigure}
drawn from our \textbf{unfiltered} data, $(T_{\text{inst}}, I_{\text{person}}, I^{\text{edit}}_{\text{person}})$, by choosing to either keep or discard each sample. We consider "good" an MLLM score $>80$ and ``bad'' otherwise, as reported in the main paper.

Based on 221 collected samples, we computed the confusion matrix between human annotations and MLLM predictions (Fig.~\ref{fig:confusion_matrix}). The results reveal a strong correlation between the fine-tuned MLLM and user preferences across both high- and low-quality samples. Notably, 77.7\% of the images were classified as ``good'' by both humans and InternVL, while 17.9\% were mutually identified as ``bad'', demonstrating the model's ability to identify noise in the raw data. In total, the MLLM achieved a 95.6\% accuracy rate in replicating human judgment, validating its efficacy as a robust filtering tool.

\subsection{Template-Based Instruction Generation}
Our editing instructions are generated using rule-based templates grounded in the structured garment attributes extracted by the MLLM. While template-driven synthesis limits the linguistic variety compared to free-form human annotations and may introduce more regular phrasing, this structure is intentional. Instruction-driven fashion editing benefits from inputs that are precise, unambiguous, and tightly coupled to garment semantics; templates ensure that each instruction directly reflects a well-defined attribute change rather than a loosely worded or ambiguous request.
More importantly, the template space spans a broad range of appearance and structural edits, and can be extended with minimal effort to support additional modification types or linguistic variations.

\subsection{Limitation and Future Work}
\label{supp:limit}
Our work establishes a strong baseline for instruction-driven fashion editing; however, certain design choices open exciting avenues for future exploration.

\tit{Data Curation and Domain Scalability}
To overcome the prohibitive cost and scarcity of manually annotated editing pairs, \dataset prioritizes scalability through an automated synthesis pipeline. While this design choice enables the training of data-hungry diffusion models, the resulting distribution is bounded by the generative capabilities of the underlying teacher models (FLUX.2 Klein~\cite{flux-2-2025} and FitDiT~\cite{jiang2024fitdit}). We mitigate potential domain artifacts through our rigorous MLLM-based verification pipeline (filtering samples with a score lower than 80), ensuring high visual fidelity. Future works could further bridge the gap between synthetic and in-the-wild distributions by incorporating small-scale, real-world human feedback loops or few-shot adaptation on physical garment pairs.

\tit{Category Expansion}
Our framework focuses on three primary garment categories (dresses, upper-body, and lower-body), which constitute the vast majority of e-commerce catalogs. While the proposed architecture is category-agnostic, extending the instruction taxonomy to handle highly occluded accessories (\eg, scarves, jewelry) or complex multi-layer styling (\eg, jackets over hoodies) remains a distinct challenge. We view this as a natural next step for the field, building upon the foundational editing capabilities established in this work.

\end{document}